\documentclass{article}

\usepackage[final,nonatbib]{neurips_2020}

\usepackage[utf8]{inputenc} 
\usepackage[T1]{fontenc}    
\usepackage{hyperref}       
\usepackage{url}            
\usepackage{booktabs}       
\usepackage{amsfonts}       
\usepackage{nicefrac}       
\usepackage{microtype}      

\usepackage[utf8]{inputenc}
\usepackage{graphicx}
\usepackage{color}
\usepackage{times}
\usepackage{amsmath,mathtools}
\usepackage{amssymb}
\usepackage{enumerate}
\usepackage{subfigure}
\usepackage[algo2e,inoutnumbered,linesnumbered,algoruled,vlined]{algorithm2e}
\usepackage{algpseudocode}
\usepackage{booktabs}
\usepackage{tabularx}
\usepackage{array}
\usepackage{bm}
\usepackage{enumitem}
\usepackage{xcolor}
\usepackage{wrapfig}
\usepackage{tabularx}
\usepackage{cutwin}

\usepackage{cleveref}

\usepackage{amsmath}
\usepackage{amssymb}
\usepackage{booktabs}
\usepackage{capt-of} 
\usepackage{cuted} 
\usepackage{etoolbox} 
\usepackage{float}
\usepackage{fontawesome}
\usepackage{graphicx}
\usepackage{makecell}
\usepackage{marvosym}
\usepackage{multirow}
\usepackage{pifont}
\usepackage{tabularx}
\usepackage{wrapfig}

\newcommand{\parahead}[1]{\noindent\textbf{#1}:\ }

\newcommand{\filluptopage}[1]{%
  \clearpage
  \loop\ifnum\value{page}<#1\relax
    \null\clearpage
  \repeat
  \loop\ifnum\value{page}=#1\relax
    \null\clearpage
  \repeat
}

\makeatletter
\def\blfootnote{\xdef\@thefnmark{}\@footnotetext}
\makeatother

\usepackage{amsmath,amsfonts,bm}

\def\eqref#1{equation~\ref{#1}}

\def\1{\bm{1}}

\DeclareMathAlphabet{\mathsfit}{\encodingdefault}{\sfdefault}{m}{sl}
\SetMathAlphabet{\mathsfit}{bold}{\encodingdefault}{\sfdefault}{bx}{n}

\newcommand{\R}{\mathbb{R}}

\let\tr\relax

\DeclareMathOperator{\tr}{Tr}

\newcommand{\x}{\mathbf{x}}
\newcommand{\z}{\mathbf{z}}
 
\newcommand{\Xset}{{\cal X}} 
 
\newcommand{\Pset}{{\cal P}} 
\newcommand{\Zset}{{\cal Z}} 
\newcommand{\Traj}{{\cal T}} 
\newcommand{\Man}{\mathcal{M}}
\newcommand{\Gauss}{\mathcal{N}}
\newcommand{\Base}{\mathcal{B}}

\newcommand{\CNFpars}{\bm{\beta}}
\newcommand{\Cpars}{\bm{\alpha}}
\newcommand{\ODEpars}{\bm{\theta}}

\newcommand{\TNOCS}{\widehat{\Pset}}
\newcommand{\GTNOCS}{\overbar{\Pset}}
\newcommand{\tnocs}{\widehat{\mathbf{p}}}
\newcommand{\gtnocs}{\overbar{\mathbf{p}}}

\newcommand{\Id}{\mathbf{I}}
\newcommand{\T}{\mathcal{T}}

\newcommand{\zero}{\mathbf{0}}

\newcommand{\y}{\mathbf{y}}

\newcommand{\overbar}[1]{\mkern 1.5mu\overline{\mkern-1.5mu#1\mkern-1.5mu}\mkern 1.5mu}

\newtheorem{dfn}{Definition}

\newcommand{\etal}{\textit{et al}. }
\newcommand{\ie}{\textit{i}.\textit{e}.}
\newcommand{\eg}{\textit{e}.\textit{g}.}
\newcommand{\cf}{\textit{c}.\textit{f}.}

\usepackage{cleveref}

\Crefname{assumption}{\textbf{H}\hspace{-3pt}}{\textbf{H}\hspace{-3pt}}
\crefname{algorithm}{\text{Alg.}}{\text{Alg.}}
\crefname{assumption}{\textbf{H}}{\textbf{H}}
\crefname{equation}{\text{Eq}}{\text{Eq}}
\crefname{definition}{\text{Dfn.}}{\text{Dfn.}}
\crefname{lemma}{\text{Lemma}}{\text{Lemma}}
\crefname{dfn}{\text{Dfn.}}{\text{Dfn.}}
\crefname{thm}{\text{Thm.}}{\text{Thm.}}
\crefname{tab}{\text{Tab.}}{\text{Tab.}}
\crefname{fig}{\text{Fig.}}{\text{Fig.}}
\crefname{table}{\text{Tab.}}{\text{Tab.}}
\crefname{figure}{\text{Fig.}}{\text{Fig.}}
\crefname{section}{\text{Sec.}}{\text{Sec.}}

\title{CaSPR: Learning {Ca}nonical {S}patiotemporal\\{P}oint Cloud {R}epresentations}

\author{Davis Rempe\textsuperscript{1} \quad Tolga Birdal\textsuperscript{1} \quad Yongheng Zhao\textsuperscript{2} \quad Zan Gojcic\textsuperscript{3} \\ \textbf{Srinath Sridhar\textsuperscript{4}}\footnote{}
 \quad \textbf{Leonidas J.~Guibas\textsuperscript{1}}\\
  \textsuperscript{1}Stanford University \quad  \textsuperscript{2}University of Padova \quad \textsuperscript{3}ETH Z\"{u}rich \quad \textsuperscript{4}Brown University \\
\href{https://geometry.stanford.edu/projects/caspr}{\texttt{geometry.stanford.edu/projects/caspr}}
}

\begin{document}
\blfootnote{\textsuperscript{*}Work done while at Stanford}
\maketitle
\vspace{-5mm}
\begin{abstract}
We propose CaSPR, a method to learn object-centric \textbf{Ca}nonical \textbf{S}patiotemporal \textbf{P}oint Cloud \textbf{R}epresentations of dynamically moving or evolving objects.
Our goal is to enable information aggregation over time and the interrogation of object state at any spatiotemporal neighborhood in the past, observed or not.
Different from previous work, CaSPR learns representations that support spacetime continuity, are robust to variable and irregularly spacetime-sampled point clouds, and generalize to unseen object instances. Our approach divides the problem into two subtasks. First, we explicitly encode time by mapping an input point cloud sequence to a spatiotemporally-canonicalized object space. We then leverage this canonicalization to learn a spatiotemporal latent representation using neural ordinary differential equations and a generative model of dynamically evolving shapes using continuous normalizing flows. We demonstrate the effectiveness of our method on several applications including shape reconstruction, camera pose estimation, continuous spatiotemporal sequence reconstruction, {and correspondence estimation} from irregularly or intermittently sampled observations.
\end{abstract}


\vspace{-3mm}
\section{Introduction}
\label{sec:intro}
\begin{wrapfigure}{r}{0.41\textwidth}
    \vspace{-15mm}
\includegraphics[width=0.41\textwidth]{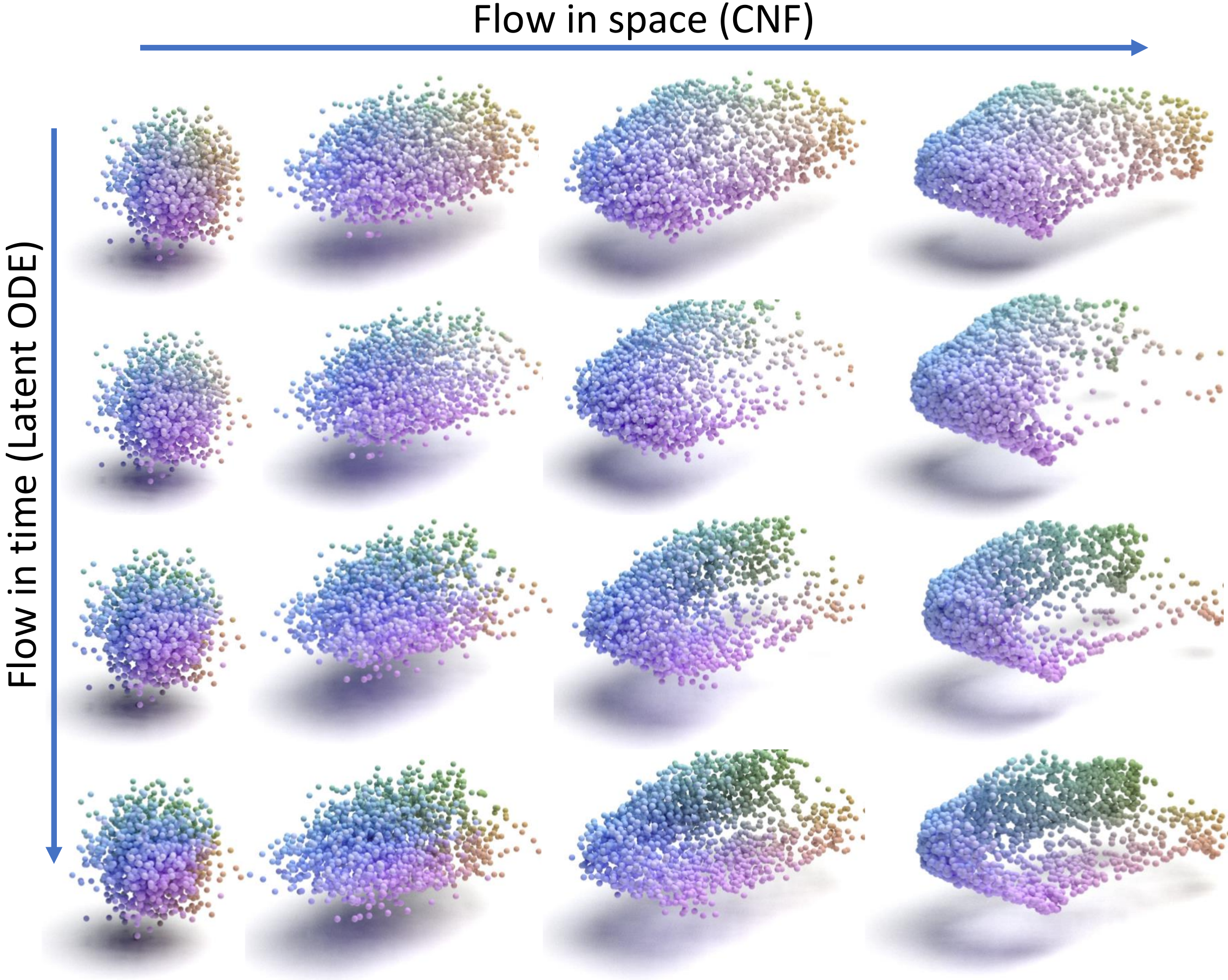}\vspace{-6mm}
    \caption{\small CaSPR builds a point cloud representation of (partially observed) objects continuously in both space ($x$-axis) and time ($y$-axis), while canonicalizing for extrinsic object properties like pose.}\vspace{-3mm}
    \label{fig:teaser}
    \vspace{-\baselineskip}
\end{wrapfigure}

The visible geometric properties of objects around us are constantly evolving over time due to object motion, articulation, deformation, or observer movement. Examples include the rigid \emph{motion} of cars on the road, the \emph{deformation} of clothes in the wind, and the \emph{articulation} of moving humans. The ability to capture and reconstruct these spatiotemporally changing geometric object properties is critical in applications like autonomous driving, robotics, and mixed reality.
Recent work has made progress on learning object shape representations from static 3D observations~\cite{deepsdf2019,qi2017pointnet,qi2017pointnet++,sitzmann2019deepvoxels,Wang:2018:pixel2mesh} and dynamic point clouds~\cite{chen2018neural,choy20194d,liu2019flownet3d,liu2019meteornet,niemeyer2019occupancy,Prantl2020Tranquil,zhang2019cloudlstm}. Yet, important limitations remain in terms of the lack of temporal continuity, robustness, and category-level generalization.

In this paper, we address the problem of learning object-centric representations that can aggregate and encode \textbf{spatiotemporal (ST) changes} in object shape as seen from a 3D sensor.
This is challenging since dynamic point clouds captured by depth sensors or LIDAR are often incomplete and sparsely sampled over space and time. Furthermore, even point clouds corresponding to adjacent frames in a sequence will experience large sampling variation.
Ideally, we would like spatiotemporal representations to satisfy several desirable properties.
First, representations should allow us to capture object shape \textbf{continuously} over space \emph{and} time.
They should encode changes in shape due to varying camera pose or temporal dynamics,
and support shape generation at arbitrary spatiotemporal resolutions.
Second, representations should be \textbf{robust} to irregular sampling patterns in space and time, including support for full or partial point clouds.
Finally, representations should support within-category \textbf{generalization} to unseen object instances and to unseen temporal dynamics.
While many of these properties are individually considered in prior work~\cite{choy20194d,kanazawa2019learning,liu2019meteornet,niemeyer2019occupancy,varanasi2008temporal}, a unified and rigorous treatment of all these factors in space and time is largely missing.

We address the limitations of previous work by learning a novel object-centric ST representation which satisfies the above properties. To this end, we introduce \textbf{CaSPR} -- a method to learn \textbf{Ca}nonical \textbf{S}patiotemporal \textbf{P}oint Cloud \textbf{R}epresentations.
In our approach, we split the task into two: (1)~\emph{canonicalizing} an input object point cloud sequence (partial or complete) into a shared 4D container space, and (2)~learning a continuous ST latent representation on top of this canonicalized space. 
For the former, we build upon the Normalized Object Coordinate Space (NOCS)~\cite{xnocs_sridhar2019,wang2019normalized} which canonicalizes intra-class 3D shape variation by normalizing for extrinsic properties like position, orientation, and scale.
We extend NOCS to a 4D \textbf{Temporal-NOCS (T-NOCS)}, which additionally normalizes the duration of the input sequence to a unit interval.
Given dynamic point cloud sequences, our ST canonicalization yields spacetime-normalized point clouds.
In~\cref{sec:exp}, we show that this allows learning representations that generalize to novel shapes and dynamics.

We learn ST representations of canonicalized point clouds using \emph{Neural Ordinary Differential Equations} (Neural ODEs)~\cite{chen2018neural}.
Different from previous work, we use a \textbf{Latent ODE} that operates in a lower-dimensional \emph{learned latent space} which increases efficiency while still capturing object shape dynamics.
Given an input sequence, the canonicalization network and Latent ODE together extract features that constitute an ST representation.
To continuously \emph{generate} novel spatiotemporal point clouds conditioned on an input sequence, we further leverage invertible \emph{Continuous Normalizing Flows} (CNFs)~\cite{Chen2018,grathwohl2019ffjord} which transform Gaussian noise directly to the visible part of an object's shape at a desired timestep.
Besides continuity, CNFs provide direct likelihood evaluation which we use as a training loss.
Together, as shown in~\cref{fig:teaser}, the Latent ODE and CNF constitute a generative model that is continuous in spacetime and robust to sparse and varied inputs.
Unlike previous work~\cite{choy20194d,liu2019meteornet}, our approach is continuous and explicitly avoids treating time as 
another spatial dimension by respecting its unique aspects (\eg, unidirectionality).
    
We demonstrate that CaSPR is useful in numerous applications including (1)~continuous spacetime shape reconstruction from sparse, partial, or temporally non-uniform input point cloud sequences, (2)~spatiotemporal 6D pose estimation, and (3)~information propagation via space-time correspondences under rigid or non-rigid transformations.
Our experiments show improvements to previous work while also providing insights on the emergence of intra-class shape correspondence and the learning of \emph{time unidirectionality}~\cite{eddington2019nature}.
In summary, our contributions are:
\begin{enumerate}[itemsep=0pt,topsep=0pt,leftmargin=*]
\item The CaSPR encoder network that consumes dynamic object point cloud sequences and canonicalizes them to normalized spacetime (T-NOCS).
\item The CaSPR representation of canonicalized point clouds using a Latent ODE to explicitly encode temporal dynamics, and an associated CNF for generating shapes continuously in spacetime.
\item A diverse set of applications of this technique, including partial or full shape reconstruction, spatiotemporal sequence recovery, camera pose estimation, and correspondence estimation.
\end{enumerate}
\section{Related Work}
\label{sec:related}\vspace{-2mm}
\paragraph{Neural Representations of Point Sets} 
Advances in 2D deep architectures leapt into the realm of point clouds with PointNet~\cite{qi2017pointnet}. The lack of locality in PointNet was later addressed by a diverse set of works~\cite{deng2018eccv,deng2018ppfnet,li2018pointcnn,shen2018mining,su2018splatnet,thomas2019kpconv,wang2019dynamic,yang2020continuous,zhao20193d}, including  PointNet++~\cite{qi2017pointnet++} -- a permutation invariant architecture capable of learning both local and global point features. We refer the reader to Guo~\etal~\cite{guo2019deep} for a thorough review. Treating time as the fourth dimension, our method heavily leverages propositions from these works. 
Continuous reconstruction of an object's spatial geometry has been explored by recent works in learning implicit shape representations~\cite{chen2019learning,zekun2020dualsdf,mescheder2019occupancy,deepsdf2019}.

\vspace{-2mm}
\paragraph{Spatiotemporal Networks for 3D Data}
Analogous to volumetric 3D convolutions on video frames~\cite{lea2017temporal,varol2017long,Zhang2020V4D}, a direct way to process spatiotemporal point cloud data is performing 4D convolutions on a voxel representation. This poses three challenges: (1) storing 4D volumes densely is inefficient and impractical, (2) direct correlation of spatial and temporal distances is undesirable, and (3) the inability to account for timestamps can hinder the final performance. These challenges have fostered further research along multiple fronts.
For example, a large body of works~\cite{behl2019pointflownet,gu2019hplflownet,liu2019flownet3d,wang2019flownet3d++} has addressed temporal changes between a pair of scans as per-point displacements or \textit{scene flow}~\cite{vedula1999three}.
While representing dynamics as fields of change is tempting, such methods lack an explicit notion of time.
MeteorNet~\cite{liu2019meteornet} was an early work to learn flow on raw point cloud sequences, however it requires explicit local ST neighborhoods which is undesirable for accuracy and generalization. Prant~\etal~\cite{Prantl2020Tranquil} use temporal frames as a cue of coherence to stabilize the generation of points.
CloudLSTM~\cite{zhang2019cloudlstm} models temporal dependencies implicitly within sequence-to-sequence learning. Making use of time in a more direct fashion, MinkowskiNet~\cite{choy20194d} proposed an efficient ST 4D CNN to exploit the sparsity of point sets. This method can efficiently perform 4D sparse convolutions, but can neither canonicalize time nor perform ST aggregation. OccupancyFlow~\cite{niemeyer2019occupancy} used occupancy networks~\cite{mescheder2019occupancy} and Neural ODEs~\cite{chen2018neural} to have an explicit notion of time.

Our method can be viewed as learning the underlying \emph{kinematic spacetime surface} of an object motion: an idea from traditional computer vision literature for dynamic geometry registration~\cite{mitra2007dynamic}.
\vspace{-2mm}
\paragraph{Canonicalization}
Regressing 3D points in a common global reference frame dates back to 6D camera relocalization and is known as \emph{scene coordinates}~\cite{shotton2013scene}. In the context of learning the \emph{normalized object coordinate space} (NOCS), \cite{wang2019normalized} is notable for explicitly mapping the input to canonical \emph{object coordinates}. Thanks to this normalization, NOCS enabled category-level pose estimation and has been extended to articulated objects~\cite{li2019category}, category-level rigid 3D reconstruction~\cite{choy20163d,groueix2018atlasNet,kar2015category} via multiview aggregation~\cite{xnocs_sridhar2019}, and non-rigid shape reconstruction either via deep implicit surfaces~\cite{zakharov2019autolabeling} or by disentangling viewpoint and deformation~\cite{novotny2019c3dpo}. Chen~\etal~\cite{chen2020learning} proposed a latent variational NOCS to \emph{generate} points in a canonical frame. 

\vspace{-2mm}
\paragraph{Normalizing Flows and Neural ODEs}
The idea of transforming noise into data dates back to whitening transforms~\cite{friedman1987exploratory} and Gaussianization~\cite{Chen2001}. Tabak and Turner~\cite{tabak2013family} officially defined normalizing flows (NFs) as the composition of simple maps and used it for non-parametric density estimation. NFs were immediately extended to deep networks and high dimensional data by Rippel and Adams~\cite{rippel2013high}. Rezende and Mohamed used NFs in the setting of variational inference~\cite{rezende2015variational} and popularized them as a standalone tool for deep generative modeling \eg~ \cite{kingma2018glow,sun2019dual}. Thanks to their invertibility and exact likelihood estimation, NFs are now prevalent and have been explored in the context of graph neural networks~\cite{liu2019graph}, generative adversarial networks~\cite{grover2018flow}, bypassing topological limitations~\cite{atanov2019semi,cornish2019relaxing,dupont2019augmented}, flows on Riemannian manifolds~\cite{gemici2016normalizing,lou2020neural,salman2018deep}, equivariant flows~\cite{bilovs2020equivariant,kohler2019equivariant,rezende2019equivariant}, and connections to optimal transport~\cite{finlay2020train,onken2020ot,villani2008optimal,zhang2019mongeampere}.
The limit case where the sequence of transformations are indexed by real numbers yields continuous-time flows: the celebrated Neural ODEs~\cite{Chen2018}, their latent counterparts~\cite{rubanova2019latent}, and 
FFJORD~\cite{grathwohl2019ffjord}, an invertible generative model with unbiased density estimation. For a comprehensive review, we refer the reader to the concurrent surveys of~\cite{kobyzev2019normalizing,papamakarios2019normalizing}.

Our algorithm is highly connected to PointFlow~\cite{yang2019pointflow} and C-Flow~\cite{pumarola2019c}. However, we tackle encoding and generating spatiotemporal point sets in addition to canonicalization while both of these works use CNFs in generative modeling of 3D point sets without canonicalizing.
\vspace{-2mm}\section{Background}\vspace{-1mm}
\label{sec:bg}
In this section, we lay out the notation and mathematical background required in \cref{sec:method}.
\begin{dfn}[Flow \& Trajectory]
Let us define a $d$-dimensional \textbf{flow} to be a parametric family of homeomorphisms $\phi:\Man\times\R \mapsto \Man$ acting on a vector $\z\in\Man\subset\R^d$ with $\phi_0(\z)=\z$ (identity map) and $\phi_t(\z)=\z_t$.
A temporal subspace of flows is said to be a \textbf{trajectory} $\T(\z)=\{\phi_t(\z)\}_t$ if $\Traj(\z)\cap\Traj(\y)=\emptyset\,\text{ for all } \z\neq\y$, \ie, different trajectories never intersect~\cite{coddington1955theory,dupont2019augmented}.
\end{dfn}\vspace{-2mm}
\begin{dfn}[ODE-Flow, Neural ODE \& Latent ODE]\label{dfn:latentODE}
For any given flow $\phi$ there exists a corresponding ordinary differential equation (\textbf{ODE}) constructed by attaching an optionally time-dependent vector $f(\z, t)\in\R^d$ to every point $\z\in\Man$ resulting in a vector field s.t. $f(\z)= \phi^{\prime}(\z)|_{t=0}$. Starting from the initial state $\z_0$, this ODE given by $\frac{d{z(t)}}{dt} = f(z(t),t)$ can be integrated for time $T$ modeling the flow $\phi_{t=T}$:
\begin{equation}
\label{eq:ode}
    \z_T = \phi_T(\z_0) = \z_0 + \int_{0}^T f_{\theta}(\z_t, t) \, dt,
\end{equation}
where $\z_t\triangleq z(t)$ and the field $f$ is parameterized by $\ODEpars=\{\theta_i\}_i$. By the Picard–Lindelöf theorem~\cite{coddington1955theory}, if $f$ is continuously differentiable then the initial value problem in~\cref{eq:ode} has a unique solution. 
Instead of handcrafting, \textbf{Neural ODEs}~\cite{chen2018neural} seek a function $f$ that suits a given objective by modeling $f$ as a neural network.
We refer to a Neural ODE operating in a latent space as a \textbf{Latent ODE}.
\end{dfn}\vspace{-1mm}
Numerous forms of Neural ODEs model $f(\cdot)$ to be \emph{autonomous}, \ie, time independent $f(\z_t)\equiv f(\z_t, t)$~\cite{chen2018neural,dupont2019augmented,rubanova2019latent}, whose output fully characterizes the trajectory.
While a Neural ODE advects single particles, \textbf{generative modeling} approximates the full target probability density which requires expressive models capable of exact density evaluation and sampling that avoids mode collapse.
\begin{dfn}[Continuous Normalizing Flow (CNF)]
Starting from a simple $d_{\Base}$-dimensional base distribution $p_y$ with $\y_0\in\R^{d_{\Base}}\sim p_y(\y)$, \textbf{CNF}s~\cite{Chen2018,grathwohl2019ffjord} aim to approximate the complex target distribution $p_{x}(\x)$ by bijectively mapping empirical samples of the target to the base using an invertible function $g_{\beta}:\R^{d_{\Base}}\mapsto\R^{d_{\Base}}$ with parameters $\CNFpars=\{\beta_i\}_i$.
Then the probability density function transforms with respect to the change of variables: $\log p_{x}(\x) = \log p_y(\y) - \log\det\nabla g_\beta(\y)$. 
The warping function $g$ can be replaced by an integral of continuous-time dynamics yielding a form similar to Neural ODEs except that we now consider distributions~\cite{grathwohl2019ffjord}:
\begin{equation}\label{eq:condcnf}
\log p_{x}(\x) \,=\, \log p_y(\y_0) - \int_{0}^T \tr\Big(\frac{\partial g_{\beta}(\y_t,t\,|\, \z)}{\partial \y_t} \Big) \, dt,
\end{equation}
with the simplest choice that the base distribution $\y_0$ is in a $d$-dimensional ball, $p_y\sim \Gauss(\zero,\Id)$. Here $\z\in\R^d$ is an optional \textbf{conditioning} latent vector~\cite{yang2019pointflow}. Note that this continuous system is \textnormal{non-autonomous} \ie, time varying and every non-autonomous system can be converted to an autonomous one by raising the dimension to include time~\cite{davis2020time,dupont2019augmented}.
\end{dfn}\vspace{-2mm}
\vspace{-2mm}\section{Method}
\begin{figure*}[t!]
    \centering
  \includegraphics[width=\textwidth]{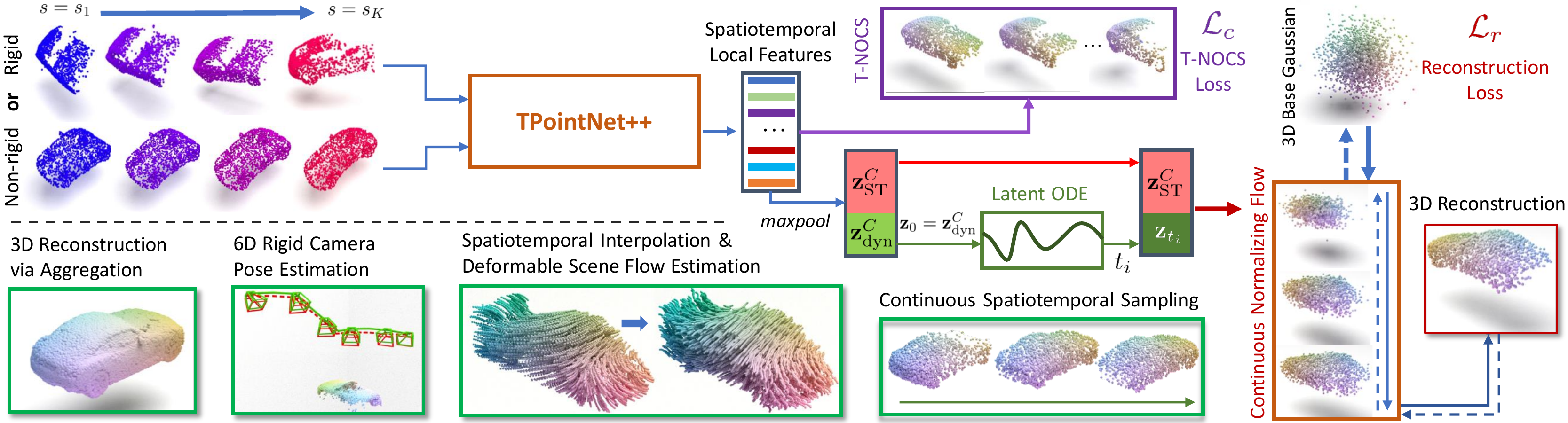}\vspace{-2mm}
    \caption{\small Architecture and applications of CaSPR. Our model consumes rigid or deformable point cloud sequences and maps them to a spatiotemporal canonical latent space whose coordinates are visualized by RGB colors (purple box).
    Using a Latent ODE, it advects a latent subspace forward in time to model temporal dynamics. A continuous normalizing flow~\cite{grathwohl2019ffjord} (shown in red) decodes the final latent code to 3D space by mapping Gaussian noise to the partial or full shape at desired timesteps. CaSPR enables multiple applications shown in green boxes. Training directions for the normalizing flow are indicated by dashed arrows.\vspace{-5mm}
    }
    \label{fig:samplingarch}
\end{figure*}

\vspace{-1mm}
\label{sec:method}
We consider as input a sequence of potentially partial, clutter-free 3D scans (readily captured by depth sensors or LIDAR) of an object belonging to a known category.
This observation is represented as a point cloud $\Xset=\{\mathbf{x}_i\in\R^3=\{x_i,y_i,z_i\} \,|\,i=1,\dots,M^{\prime}\}$.
For a sequence of $K$ potentially non-uniformly sampled timesteps, we denote a
\textbf{spatiotemporal} (ST) point cloud as $\Pset=\{\Pset_k\}_{k=1}^K$, where $\Pset_k = \{\mathbf{p}_i\in\R^4=\{x_i,y_i,z_i,s_k\} \,|\, i = 1,\dots,M_k\}$, $M_k$ is the number of points at frame $k\in[1,K]$ and at the \emph{time} $s_k\in[s_1,s_K]\subset\R$ with $M=\sum_{k=1}^K M_k$. Our goal is to explain $\Pset$ by learning a \emph{continuous representation} of shape that is invariant to extrinsic properties while aggregating intrinsic properties along the direction of time.
\textbf{CaSPR} achieves this through:
\begin{enumerate}[itemsep=0pt,topsep=0pt,leftmargin=*]
    \item~A \textbf{canonical} spacetime container where extrinsic properties such as object pose are factored out,
    \item~A continuous \textbf{latent} representation which can be queried at arbitrary spacetime steps, and
    \item~A \textbf{generative} model capable of reconstructing partial observations conditioned on a latent code.
\end{enumerate}
We first describe the method design for each of these components, depicted in~\cref{fig:samplingarch}, followed by implementation and architectural details in~\cref{sec:network}.

\begin{figure*}[t!]
    \centering
  \includegraphics[width=\textwidth]{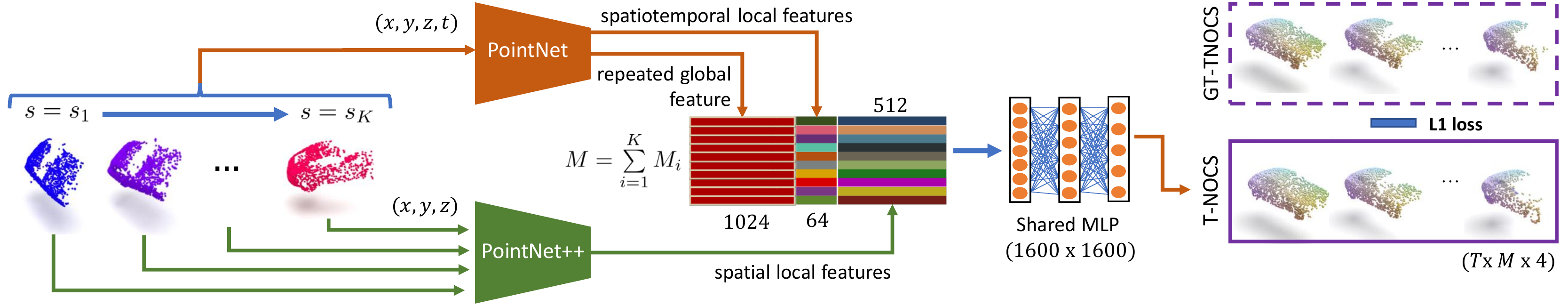}\vspace{-3mm}
    \caption{\small Architecture of our ST point-set canonicalization network, \emph{TPointNet++}. It uses two branches that extract ST features using a 4D PointNet and per-view 3D local features via PointNet++. These features are combined and passed to an MLP to regress the T-NOCS points. Training is supervised via GT coordinates.\vspace{-7mm}}
    \label{fig:pointnetplus}
\end{figure*}

\parahead{Canonicalization} 
The first step is \emph{canonicalization} of a 4D ST point cloud sequence with the goal of associating observations at different time steps to a common canonical space.
Unlike prior work which assumes already-canonical inputs~\cite{niemeyer2019occupancy,yang2019pointflow}, this step allows CaSPR to operate on raw point cloud sequences in world space and enables multiple applications (see~\cref{fig:samplingarch}).
Other previous work has considered canonicalization of extrinsic properties from RGB images~\cite{xnocs_sridhar2019,wang2019normalized} or a 3D point cloud~\cite{li2019category}, but our method operates on a 4D point cloud and explicitly accounts for time labels.
Our goal is to find an injective \emph{spacetime canonicalizer} $c_\alpha(\cdot)\, :\, \Pset \mapsto \GTNOCS \times \overbar{\Zset}$ parameterized by $\Cpars=\{\alpha_i\}_i$, that maps a point cloud sequence $\Pset$ to a canonical \emph{unit tesseract} $\GTNOCS = \{\GTNOCS_k\}_{k=1}^K$, where $\GTNOCS_k=\{\mathbf{\overbar{p}}_i\in\R^4=\{\overbar{x}_i,\overbar{y}_i,\overbar{z}_i,\overbar{t}_k\}\in [0,1] \,|\, i = 1,\dots,M_k\}$ and $\z^C\in\overbar{\Zset}\subset\R^d$ is the corresponding canonical \emph{latent} representation (embedding) of the sequence.
Note that in addition to position and orientation, $\GTNOCS$ is normalized to have time in unit duration.
We refer to $\GTNOCS$ as \textbf{Temporal-NOCS (T-NOCS)} as it extends NOCS~\cite{xnocs_sridhar2019,wang2019normalized}. T-NOCS points are visualized using the spatial coordinate as the \emph{RGB} color in~\cref{fig:samplingarch,fig:pointnetplus}.
Given a 4D point cloud in the world frame, we can aggregate the entire shape from $K$ partial views by a simple union: $\GTNOCS=\bigcup_{i=1}^K c_\alpha(\Pset_i)$~\cite{xnocs_sridhar2019}.
Moreover, due to its injectivity, $c_\alpha(\cdot)$ preserves \emph{correspondences}, a property useful in tasks like pose estimation or label propagation. 
We outline the details and challenges involved in designing a canonicalizer in~\cref{sec:network}.

\parahead{Continuous Spatiotemporal Representation}
While a global ST latent embedding is beneficial for canonicalization and aggregation of partial point clouds, we are interested in continuously modeling the ST input, \ie, being able to compute a representation for unobserved timesteps at arbitrary spacetime resolutions.
To achieve this, we split the latent representation: $\z^C=[\z^C_{\text{ST}},\,\z^C_{\text{dyn}}]$ where $\z^C_{\text{ST}}$ is the \emph{static} ST descriptor and $\z^C_{\text{dyn}}$ is used to initialize an autonomous Latent ODE $\frac{d \z_t}{dt} = f_{\theta}(\z_t)$ as described in~\cref{dfn:latentODE}: $\z_0=\z^C_{\text{dyn}}\in\R^d$. 
We choose to advect the ODE in the latent space (rather than physical space~\cite{niemeyer2019occupancy}) to (1) enable \emph{learning} a space best-suited to modeling the dynamics of the observed data, and (2) improve scalability due to the fixed feature size.
Due to the time-independence of $f_{\theta}$, $\z_0$ fully characterizes the latent trajectory.
Advecting $\z_0$ forward in time by solving this ODE until any canonical timestamp $T\leq 1$ yields a continuous representation in time $\z_{T}$ that can explain changing object properties.
We finally obtain a dynamic spatiotemporal representation in the product space: $\z\in\R^{D} = [\z^C_{\text{ST}},\z_{T}]$. Due to canonicalization to the unit interval, $T>1$ implies \emph{extrapolation}.

\parahead{Spatiotemporal Generative Model}
Numerous methods exist for point set generation~\cite{achlioptas2017latent_pc,groueix2018atlasNet,zhao20193d}, but most are not suited for sampling on the surface of a partial 4D ST point cloud.
Therefore, we adapt CNFs~\cite{grathwohl2019ffjord,yang2019pointflow} as defined in~\cref{sec:bg}. To generate a novel ST shape, \ie, a sequence of 3D shapes $\Xset_1\dots \Xset_K$, we simulate the Latent ODE for $t=0\dots T$ and obtain representations for each of the canonical shapes in the sequence: $\z_{t=0}\cdots \z_{t=T}$. We then sample the base distribution $\y_k\in\R^{d_{\Base}=3} \sim p_y(\y)\triangleq \Gauss(\zero, \Id)$ and evaluate the conditional CNF in~\cref{eq:condcnf} by passing each sample $\y_k$ through the flow $g_{\beta}(\y_k \, |\, \z_t)$ conditioned on $\z_t$. Note that the flow is time dependent, \ie, non-autonomous.
To increase the temporal resolution of the output samples we pick the timesteps with higher frequency, whereas to \emph{densify} spatially, we simply generate more samples $\y_k$.

\vspace{-2mm}\subsection{Network Architecture}\vspace{-1mm}
\label{sec:network}
We now detail our implementations of the canonicalizer $c_{\alpha}$, 
Latent ODE network $f_\theta$, and CNF $g_{\beta}$. 

\parahead{TPointNet++ $c_\alpha(\cdot)$}
The design of our canonicalizer is influenced by (1)~the desire to avoid ST neighborhood queries, (2)~to treat time as important as the spatial dimensions, and (3)~injecting how an object appears during motion in space into its local descriptors resulting in more expressive features.
While it is tempting to directly apply existing point cloud architectures such as PointNet~\cite{qi2017pointnet} or PointNet++~\cite{qi2017pointnet++}, we found experimentally that they were individually insufficient (\cf~\cref{sec:exp}).
To meet our goals, we instead introduce a hybrid \emph{TPointNet++} architecture as shown in~\cref{fig:pointnetplus} to implement $c_\alpha$ and canonicalize $\Pset$ to $\GTNOCS$.
TPointNet++ contains a PointNet branch that consumes the entire 4D point cloud to extract both a 1024-dimensional global feature and 64-dimensional per-point ST features.
This treats time explicitly and equally to each spatial dimension.
We also use PointNet++ to extract a 512-dimensional local feature at each input point 
by applying it at each cross-section in time with no timestamp.
We feed all features into a shared multi-layer perceptron (MLP) to arrive at 1600-dimensional embeddings
corresponding to each input point.

We use the pointwise embeddings in two ways: (1) they are passed through a shared linear layer followed by a \emph{sigmoid} function to estimate the T-NOCS coordinates $\TNOCS$ which approximate the ground truth $\GTNOCS$, and 
(2) we max-pool all per-point features into a single latent representation of T-NOCS $\z^C\in\R^{1600}$ which is used by the Latent ODE and CNF as described below. The full canonicalizer $c_\alpha(\cdot)$
can be trained independently for T-NOCS regression, or jointly with a downstream task.

\parahead{Latent ODE $f_{\theta}(\cdot)$ and Reconstruction CNF $g_{\beta}(\cdot)$}
The full CaSPR architecture is depicted in~\cref{fig:samplingarch}. It builds upon the embedding from TPointNet++ by first splitting it into two parts $\z^C=[\z^C_{\text{ST}},\,\z_0\triangleq\z^C_{\text{dyn}}]$. The dynamics network of the Latent ODE $f_\theta$ is an MLP with three hidden layers of size $512$. We use a Runge-Kutta 4(5) solver~\cite{kutta1901beitrag,runge1895numerische} with adaptive step sizes which supports backpropagation using the adjoint method~\cite{chen2018neural}. The static feature, $\z^C_{\text{ST}}\in\R^{1536}$ is skip-connected and concatenated with $\z_{T}$ to yield $\z\in\R^{1600}$ which conditions the reconstruction at $t=T$. 

To sample the surface represented by $\z$, we use a FFJORD conditional-CNF~\cite{grathwohl2019ffjord,yang2019pointflow} as explained in~\cref{sec:bg,sec:method} to map 3D Gaussian noise $\y_0\in\R^{d_{\Base}=3} \sim \Gauss(\zero, \Id)$ onto the shape surface. The dynamics of this flow $g_{\beta}(\y_t, t \,|\, \z)$ are learned with a modified MLP~\cite{grathwohl2019ffjord} which leverages a gating mechanism at each layer to inject information about the current context including $\z$ and current time $t$ of the flow. This MLP contains three hidden layers of size $512$, and we use the same solver as the Latent ODE. Please refer to the supplement for additional architectural details.

\parahead{Training and Inference} 
CaSPR is trained with two objectives that use the GT canonical point cloud sequence $\GTNOCS$ as supervision. We primarily seek to maximize the \emph{log-likelihood} of canonical spatial points on the surface of the object when mapped to the base Gaussian using the CNF. This reconstruction loss is $\mathcal{L}_{r} = - \sum_{k=1}^K \sum_{i=1}^{M_k} \log p_x(\overbar{\mathbf{x}}_i \,|\, \z_{t_k}) $ where $\overbar{\mathbf{x}}_i$ is the spatial part of $\gtnocs_i \in \GTNOCS_k$ and the log-likelihood is computed using~\cref{eq:condcnf}. Secondly, we supervise the T-NOCS predictions from TPointNet++ with an L1 loss $\mathcal{L}_{c} = \sum_{i=1}^{M} |\tnocs_i - \gtnocs_i|$ with $\gtnocs_i \in \GTNOCS$ and $\tnocs_i \in \TNOCS$. We jointly train TPointNet++, the Latent ODE, and CNF for $\Cpars$, $\ODEpars$ and $\CNFpars$ respectively with the final loss $\mathcal{L}=\mathcal{L}_{r} + \mathcal{L}_{c}$.
During inference, TPointNet++ processes a raw point cloud sequence of an unseen shape and motion to obtain the ST embedding and canonicalized T-NOCS points. The Latent ODE, initialized by this embedding, is solved forward in time to any number of canonical ``query'' timestamps. For each timestamp, the Latent ODE produces the feature to condition the CNF which reconstructs the object surface by the forward flow of Gaussian samples. The combined continuity of the Latent ODE and CNF enables CaSPR to reconstruct the input sequence at any desired ST resolution.

\begin{figure*}[t!]
    \centering
    \includegraphics[width=\textwidth]{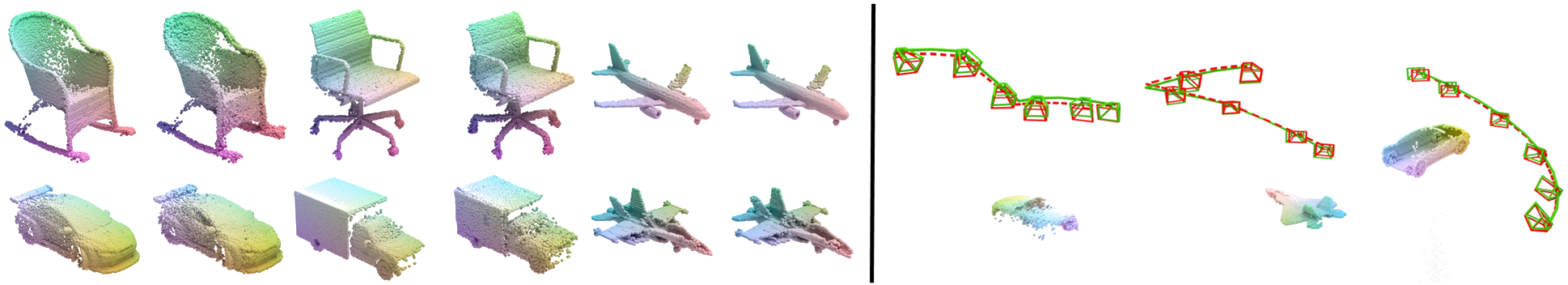}\vspace{-2mm}
    \caption{\small Canonicalization applications. Partial shape reconstruction (left section) shows pairs of GT (left) and predicted shapes (right). Pose estimation (right section) shows GT (green, solid) and predicted (red, dashed) camera pose based on regressed T-NOCS points. Points are colored by their T-NOCS location.}
    \label{fig:canonapps}
    \vspace{-1.5\baselineskip}
\end{figure*}
\vspace{-3mm}
\section{Experimental Evaluations}
\label{sec:exp}
We now evaluate the canonicalization, representation, and reconstruction capabilities of CaSPR, demonstrate its utility in multiple downstream tasks, and justify design choices.

\parahead{Dataset and Preprocessing}
We introduce a new dataset containing simulated rigid motion of objects in three ShapeNet~\cite{chang2015shapenet} categories: cars, chairs, and airplanes. The motion is produced with randomly generated camera trajectories (\cref{fig:canonapps}) and allows us to obtain the necessary inputs and supervision for CaSPR: sequences of raw partial point clouds from depth maps with corresponding canonical T-NOCS point clouds.
Each sequence contains $K=10$ frames with associated timestamps. Raw point cloud sequences are labeled with uniform timestamps from $s_1=0.0$ to $s_K=5.0$
while canonicalized timestamps range from $\overbar{t}_1 = 0$ to $\overbar{t}_K=1$.
For training, 5 frames with 1024 points are randomly subsampled from each sequence, giving non-uniform step sizes between observations.
At test time, we use a \emph{different spatiotemporal sampling} for sequences of held-out object instances: all 10 frames, each with 2048 points.
Separate CaSPR models are trained for each shape category.

\parahead{Evaluation Procedure}
To measure canonicalization errors, T-NOCS coordinates are split into the spatial and temporal part with GT given by ${\bar{\Xset}}$ and $\overbar{\mathbf{t}}$ respectively. The \emph{spatial error} at frame $k$ is $\frac{1}{M_k}\sum_{i=1}^{M_k} \left\lVert \widehat{\mathbf{x}}_i - \overbar{\mathbf{x}}_i \right\rVert_2$ and the \emph{temporal error} is $\frac{1}{M_k}\sum_{i=1}^{M_k} | \widehat{t}_i - \overbar{t_i} |$ . For reconstruction, the Chamfer Distance (CD) and Earth Mover's Distance (EMD) are measured (and reported multiplied by $10^3$). Lower is better for all metrics; we report the median over all test frames because outlier shapes cause less informative mean errors. Unless stated otherwise, qualitative point cloud results (\eg,~\cref{fig:canonapps}) are colored by
their canonical coordinate values (so corresponding points should have the same color).
\begin{wraptable}[9]{r}{0.46\textwidth}
\vspace{-14pt}
\caption{\small Canonicalization performance.}
\begin{center}
\scalebox{0.7}{
\setlength{\tabcolsep}{4pt}
\begin{tabular}{ r l c c }
\toprule
 \textbf{Method} & \textbf{Category} & \textbf{Spatial Err} & \textbf{Time Err} \\
\midrule
MeteorNet &  Cars & 0.0633 & \textbf{0.0001} \\
PointNet++ No Time &   &  0.0530 & --- \\
PointNet++ w/ Time &   &  0.0510 & 0.0005  \\
PointNet &   & 0.0250 & 0.0012 \\
TPointNet++ No Time &   &  0.0122 & --- \\
\midrule
TPointNet++ &  Cars &  \textbf{0.0118}	& 0.0011  \\
TPointNet++ &  Chairs & 0.0102 & 0.0008 \\
TPointNet++ &  Airplanes &  0.0064 & 0.0009  \\
\bottomrule
\end{tabular}}
\end{center}
\vspace{-7pt}
\label{table:canonresults}
\end{wraptable}
\vspace{-6mm}\subsection{Evaluations and Applications}
\label{sec:canonresults}
\parahead{Canonicalization}
We first evaluate the accuracy of canonicalizing raw partial point cloud sequences to T-NOCS using TPointNet++.
\cref{table:canonresults} shows median errors over all frames in the test set. The bottom part evaluates TPointNet++ on each shape category while the top compares with baselines on cars (please see supplementary for more details).
Notably, for spatial prediction, TPointNet++ outperforms variations of both PointNet~\cite{qi2017pointnet} and PointNet++~\cite{qi2017pointnet++}, along with their spatiotemporal extension MeteorNet~\cite{liu2019meteornet}. This indicates that our ST design yields more distinctive features both spatially and temporally. MeteorNet and PointNet++ (with time) achieve impressive time errors thanks to skip connections that pass the input timestamps directly towards the end of the network. Qualitative results of canonicalization are in~\cref{fig:canonapps}.
\begin{table}[t]
\begin{minipage}[b][6cm]{0.6\linewidth}
    \vspace{-6mm}
    \caption{\small Partial surface sequence reconstruction. Chamfer (CD) and Earth Mover's Distances (EMD) are multiplied by $10^3$. On the left (\emph{10 Observed}), 10 frames are given as input and all are reconstructed. On the right, 3 frames are used as input (\emph{3 Observed}), but methods also reconstruct intermediate unseen steps (\emph{7 Unobserved}).}
    \centering
    \scalebox{0.7}{
    \begin{tabular}{ l l  c c | c c c c }
            \toprule
        & \multicolumn{1}{l}{} & \multicolumn{2}{c}{\emph{10 Observed}} & \multicolumn{2}{c}{\emph{3 Observed}} & \multicolumn{2}{c}{\emph{7 Unobserved}} \\
         \textbf{Method} & \textbf{Category} & \textbf{CD} & \multicolumn{1}{c}{\textbf{EMD}} & \textbf{CD} & \textbf{EMD} & \textbf{CD} & \textbf{EMD} \\
        \midrule
        PointFlow &  Cars &  \textbf{0.454}	& 12.838 & \textbf{0.455} & 12.743 & \textbf{0.525} & 13.911 \\
        CaSPR-Atlas & &  0.492	& 19.528 & 0.540 &	22.099 & 	0.530 & 19.635 \\
        CaSPR &  & 0.566 & \textbf{10.103} & 0.590 & \textbf{11.464} & 0.584 & \textbf{11.259} \\
        \midrule
        PointFlow &  Chairs &  0.799 & 17.267 & 0.796 & 17.294 & 0.950 & 18.442 \\
        CaSPR-Atlas &  &  \textbf{0.706}	& 48.665 & 0.723 & 48.912 & 0.749 & 47.322  \\
        CaSPR &  & 0.715 & 	\textbf{13.009} & \textbf{0.681} & \textbf{13.310}	 & \textbf{0.683} & \textbf{13.564} \\
        \midrule
        PointFlow &  Airplanes & 0.251 &	9.500 & 	0.252	& 9.534 & 	0.281 &  9.814  \\
        CaSPR-Atlas & & 0.237	& 18.827 &	0.255 &	18.525 &	0.269 & 17.933 \\
        CaSPR &  &  \textbf{0.231}	& \textbf{6.026} & 	\textbf{0.215} & 	\textbf{6.144} & 	\textbf{0.216} &	\textbf{6.175} \\
        \bottomrule
    \end{tabular}
    }
    \vspace{7pt}
\label{table:reconresults}
\end{minipage}\hfill
	\begin{minipage}[b]{0.3825\linewidth}
	    \vspace{-20mm}
		\centering
        \includegraphics[width=\linewidth]{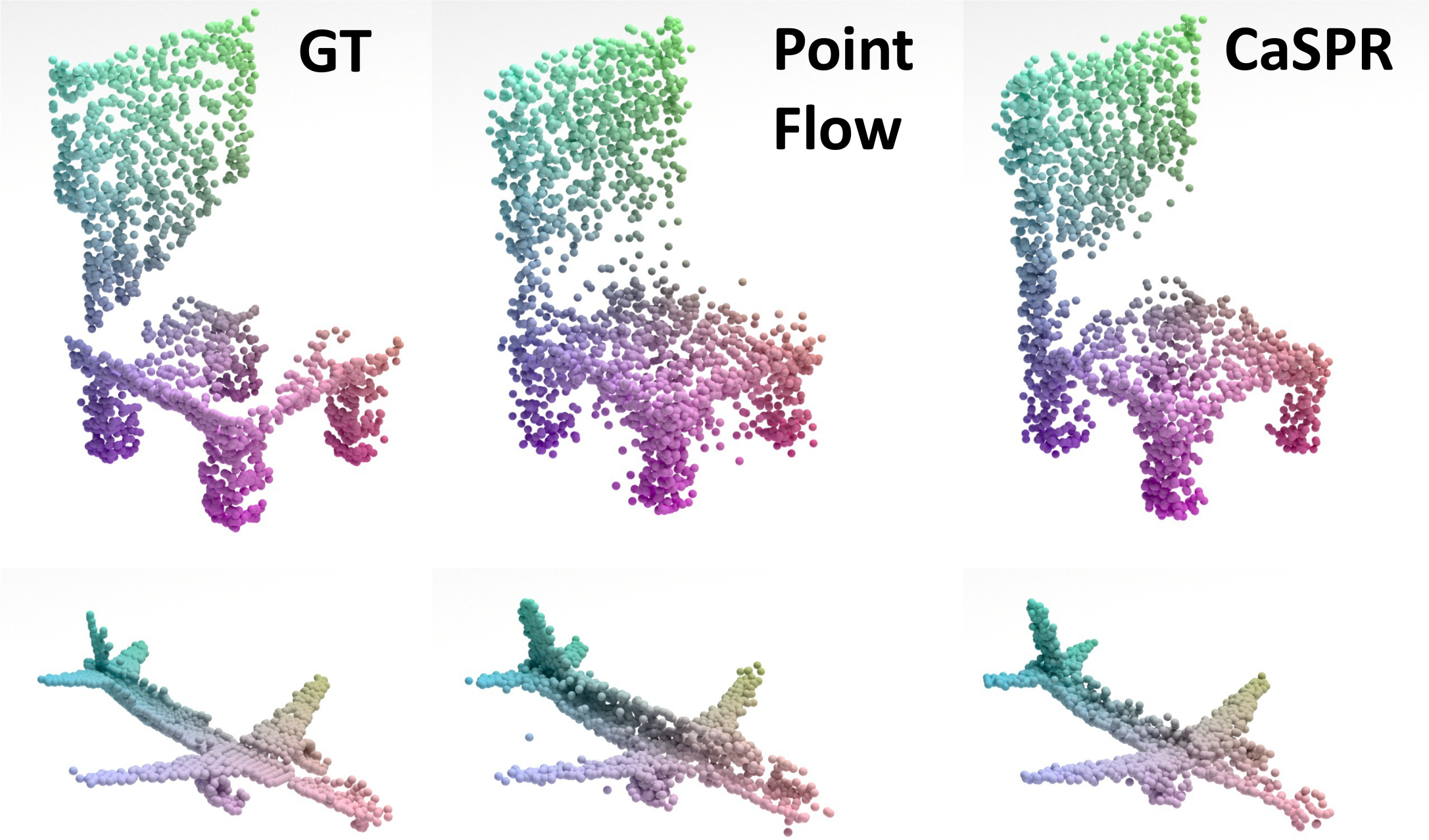}
		\captionof{figure}{\small Reconstruction results. CaSPR accurately captures occlusion boundaries for camera motion at observed and unobserved timesteps, unlike linear feature interpolation with PointFlow.}
		\label{fig:pointflowcompare}
	\end{minipage}
	\vspace{-9mm}
\end{table}

\parahead{Representation and Reconstruction}
\label{sec:reconresults}
We evaluate CaSPR's ability to represent and reconstruct observed and unobserved frames of \emph{raw} partial point cloud sequences. The full model is trained on each category separately using both $\mathcal{L}_r$ and $\mathcal{L}_c$, and is compared to two baselines. The first is a variation of CaSPR where the CNF is replaced with an AtlasNet~\cite{groueix2018atlasNet} decoder using 64 patches -- an alternative approach to achieve spatial continuity. This model is trained with $\mathcal{L}_c$ and a CD loss (rather than $\mathcal{L}_r$). The second baseline is the deterministic PointFlow~\cite{yang2019pointflow} autoencoder trained to reconstruct a \emph{single canonical} partial point cloud. This model operates on a single timestep and receives the \textbf{already canonical} point cloud as input: an easier problem. We achieve temporal continuity with PointFlow by first encoding a pair of adjacent observed point clouds to derive two shape features, and then linearly interpolating to the desired timestamp -- one alternative to attain temporal continuity. The interpolated feature conditions PointFlow's CNF to sample the partial surface, similar to CaSPR.

\cref{table:reconresults} reports median CD and EMD at reconstructed test steps for each method. We evaluate two cases: (1) models receive and reconstruct all 10 observed frames (left), and (2) models get the first, middle, and last steps of a sequence and reconstruct both these 3 observed and 7 unobserved frames (right). CaSPR outperforms PointFlow in most cases, even at observed timesteps, despite operating on raw point clouds in the world frame instead of canonical. Because PointFlow reconstructs each frame independently, it lacks temporal context resulting in degraded occlusion boundaries (\cref{fig:pointflowcompare}) and thus higher EMD. CaSPR gives consistent errors across observed and unobserved frames due to the learned motion prior of the Latent ODE, in contrast to linear feature interpolation that sees a marked performance drop for unobserved frames. The AtlasNet decoder achieves small CD since this is the primary training loss, but has difficulty reconstructing the correct point distribution on the partial surface due to the patch-based approach, resulting in much higher EMD for all cases.

\begin{wraptable}[7]{r}{0.38\textwidth}
\vspace{-27pt}
\hspace{-4mm}
\caption{\small Pose estimation using T-NOCS.}
\vspace{-3mm}
\begin{center}
\scalebox{0.65}{
\setlength{\tabcolsep}{3pt}
\hspace{-4mm}
\begin{tabular}{ l l c c c }
\toprule
\textbf{Method} & \textbf{Category} & \textbf{Trans Err} & \textbf{Rot Err}($^\circ$) & \textbf{Point Err} \\
\midrule
RPM-Net & Cars & \textbf{0.0049}  & \textbf{1.1135}  & \textbf{0.0066}  \\
CaSPR &  & 0.0077	& 1.3639 & 0.0096  \\
\midrule
RPM-Net & Chairs & \textbf{0.0026}  & \textbf{0.4601} & \textbf{0.0036}  \\
CaSPR & & 0.0075 & 1.5035 & 0.0091  \\
\midrule
RPM-Net & Airplanes & \textbf{0.0040}  & \textbf{0.5931} & \textbf{0.0048}  \\
CaSPR &  & 0.0051	& 0.9456	& 0.0057 \\
\bottomrule
\end{tabular}}
\end{center}
\vspace{3mm}
\label{table:poseresults}
\end{wraptable}
\parahead{Multiview Reconstruction}
A direct application of TPointNet++ is partial shape reconstruction of observed geometry through a union of predicted T-NOCS spatial points. Due to the quantitative accuracy of TPointNet++ at each frame (\cref{table:canonresults}), aggregated results closely match GT for unseen instances in all categories as shown in~\cref{fig:canonapps} (left).

\parahead{Rigid Pose Estimation}
The world--canonical 3D point correspondences from TPointNet++ allow fitting rigid object (or camera) pose at observed frames using RANSAC~\cite{fischler1981ransac}. \cref{table:poseresults} reports median test errors showing TPointNet++ is competitive with RPM-Net~\cite{yew2020rpmnet}, a recent \textbf{specialized} architecture for robust iterative rigid registration. Note here, RPM-Net takes \emph{both} the raw depth and \textbf{GT} T-NOCS points as input.
Translation and rotation errors are the distance and degree angle difference from the GT transformation. 
Point error measures the per frame median distance between the GT T-NOCS points transformed by the predicted pose and the input points. 
Qualitative results are in~\cref{fig:canonapps} (right).
\begin{wraptable}{r}{0.3675\textwidth}
\vspace{-20pt}
\caption{\small Reconstructing 10 observed timesteps (left) and maintaining temporal correspondences (right) on Warping Cars. }
\begin{center}
    \scalebox{0.7}{
    \begin{tabular}{ l c c | c c }
        \toprule
         & \multicolumn{2}{c}{\emph{Reconstruction}} & \multicolumn{2}{c}{\emph{Correspondences}} \\
         \textbf{Method} & \textbf{CD} & \multicolumn{1}{c}{\textbf{EMD}} & \textbf{Dist} $t_1$ & \textbf{Dist} $t_{10}$ \\
        \midrule
        OFlow &  1.512	& 20.401 & \textbf{0.011} & \textbf{0.031} \\
        CaSPR  &  \textbf{0.955}	& \textbf{11.530} & 0.013 & 0.035 \\
\bottomrule
\end{tabular}}
\end{center}
\vspace{-6mm}
\label{table:deformresults}
\end{wraptable}

\begin{figure*}[t!]
    \centering
    \includegraphics[width=\textwidth]{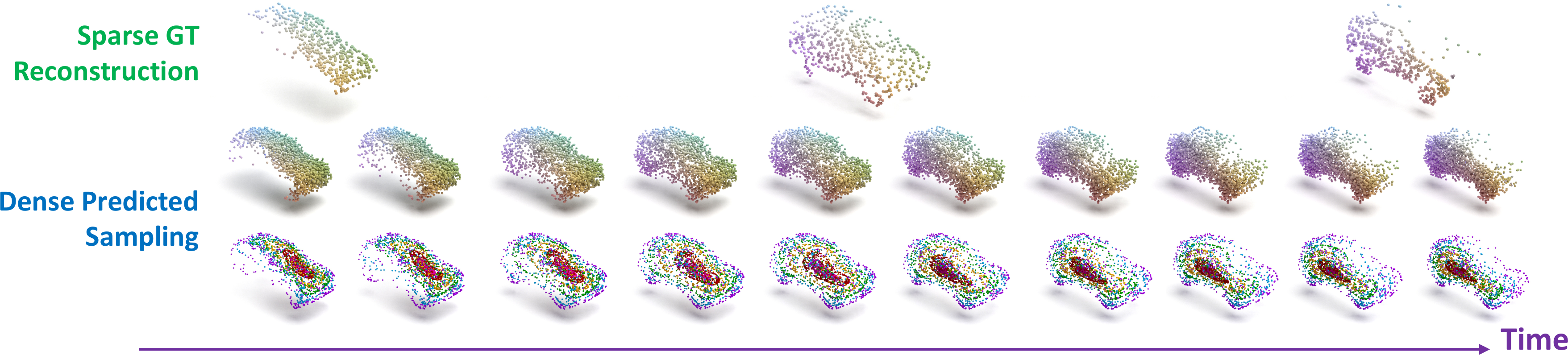}\vspace{-2mm}
    \caption{Continuous interpolation results. From three sparse frames of input with GT canonical points shown on top, CaSPR reconstructs the sequence more densely in space and time (middle). Contours of the Gaussian flowed to the car surface are shown on bottom (red is highest probability).\vspace{-3mm}}
    \label{fig:interpres}
    \vspace{-\baselineskip}
\end{figure*}

\parahead{Rigid Spatiotemporal Interpolation}
The full CaSPR model can densely sample a sparse input sequence in spacetime as shown in~\cref{fig:interpres}. The model takes three input frames of 512 points (corresponding GT T-NOCS points shown on top) and reconstructs an arbitrary number of steps with 2048 points (middle). The representation can be sampled at any ST resolution but, in practice, is limited by memory. The CNF maps Gaussian noise to the visible surface (bottom). Points are most dense in high probability areas (shown in red); in our data this roughly corresponds to where the camera is focused on the object surface at that timestep. 

\begin{wrapfigure}[14]{r}{0.2\textwidth}
    \vspace{-8mm}
\includegraphics[width=0.2\textwidth]{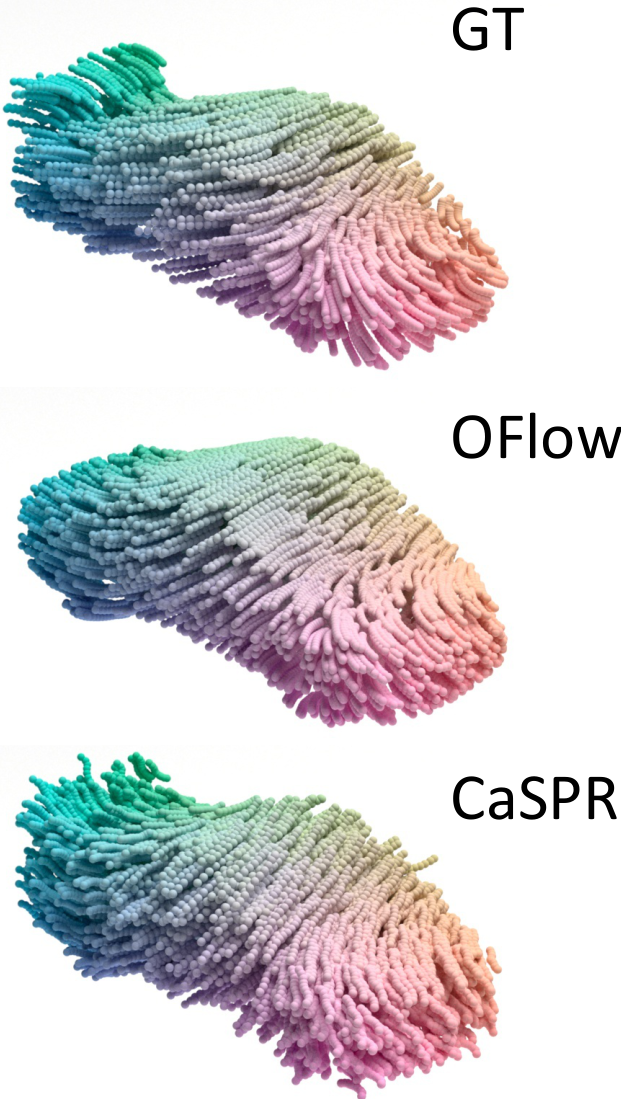}\vspace{-5mm}
    \caption{\small Deforming car reconstruction.}\vspace{-3mm}
    \label{fig:oflowcompare}
    \vspace{-3mm}
\end{wrapfigure}

\parahead{Non-Rigid Reconstruction and Temporal Correspondences}
CaSPR can represent and reconstruct deformable objects. We evaluate on a variation of the Warping Cars dataset introduced in Occupancy Flow (OFlow)~\cite{niemeyer2019occupancy} which contains 10-step sequences of \emph{full} point clouds sampled from ShapeNet~\cite{chang2015shapenet} cars deforming over time. The sequences in this dataset are already consistently aligned and scaled, so CaSPR is trained only using $\mathcal{L}_r$.

\cref{table:deformresults} compares CaSPR to OFlow on reconstructing deforming cars at 10 observed time steps (left) and on estimating correspondences over time (right). 
To measure correspondence error, we (1) sample 2048 points from the representation at $\overbar{t}_1$, (2) find their closest points on the GT mesh, and (3) advect the samples to $\overbar{t}_{10}$ and measure the mean distance to the corresponding GT points at both steps. \cref{table:deformresults} reports median errors over all $\overbar{t}_1$ and $\overbar{t}_{10}$. For OFlow, samples are advected using the predicted flow field in physical space, while for CaSPR we simply use the same Gaussian samples at each step of the sequence.

CaSPR outperforms OFlow on reconstruction due to overly-smoothed outputs from the occupancy network, while both methods accurately maintain correspondences over time. Note that CaSPR advects system state in a learned latent space and temporal correspondences \emph{naturally emerge} from the CNF when using consistent base samples across timesteps. \cref{fig:oflowcompare} visualizes sampled point trajectories for one sequence.

\begin{wrapfigure}{r}{0.25\textwidth}
    \vspace{-5mm}
\includegraphics[width=0.25\textwidth]{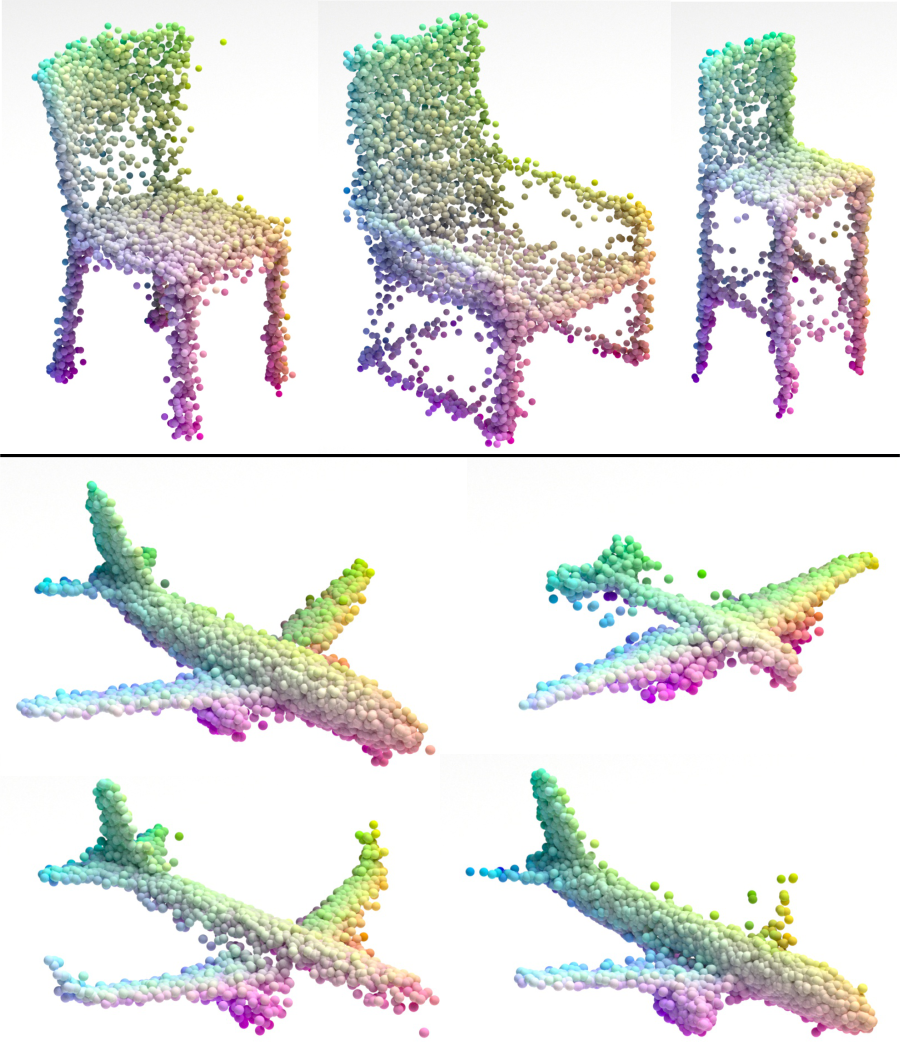}
\vspace{-4mm}
    \caption{\small Cross-instance correspondences emerge naturally using a CNF.}
    \label{fig:correspondences}
    \vspace{-5mm}
\end{wrapfigure}
\parahead{Cross-Instance Correspondences}
We observe consistent behavior from the CNF across objects within a shape category too.~\cref{fig:correspondences} shows reconstructed frames from various chair and airplane sequences with points colored by their corresponding location in the sampled Gaussian (before the flow). Similar colors across instances indicate the same part of the base distribution is mapped there. This could potentially be used, for instance, to propagate labels from known to novel object instances.

\parahead{Learning the Arrow of Time}
A desirable property of ST representations is an understanding of the \emph{unidirectionality} of time~\cite{eddington2019nature}: how objects evolve forward in time. We demonstrate this property with CaSPR by training on a dataset of 1000 sequences of a single rigid car where the camera always rotates counter-clockwise at a fixed distance (but random height). CaSPR achieves a median CD of 0.298 and \textbf{EMD of 7.114} when reconstructing held-out sequences \emph{forward in time}. 
However, when the same test sequences are \emph{reversed} by flipping the timestamps, accuracy drastically drops to CD 1.225 and \textbf{EMD 88.938}. CaSPR is sensitive to the arrow of time due to the directionality of the Latent ODE and the global temporal view provided by operating on an entire sequence jointly.

\begin{wrapfigure}{r}{0.4\textwidth}
    \vspace{-4mm}
\includegraphics[width=0.4\textwidth]{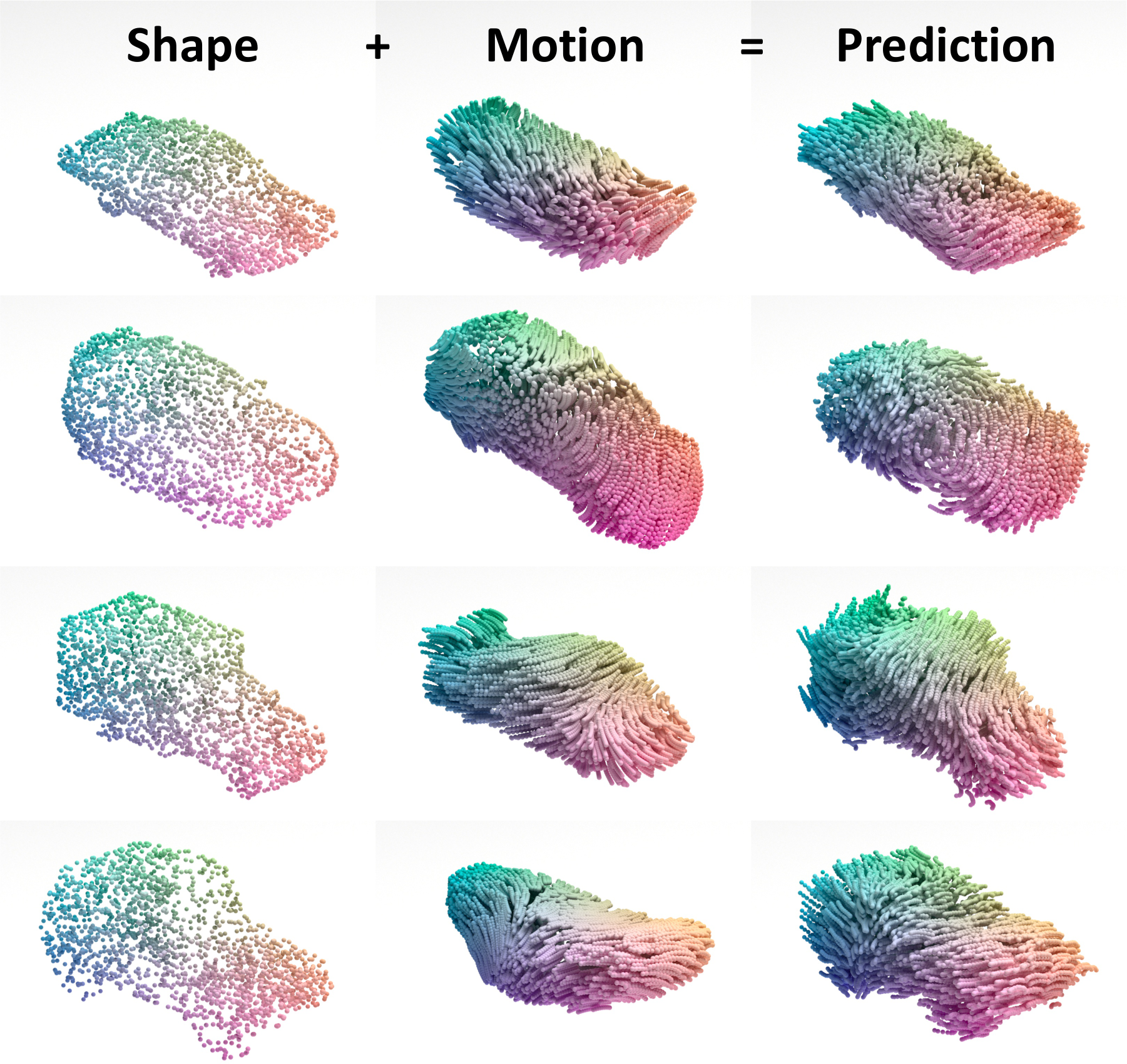}\vspace{-4mm}
    \caption{Disentanglement examples on warping cars data.}\vspace{-3mm}
    \label{fig:disentangle}
    \vspace{-3mm}
\end{wrapfigure}

\paragraph{Shape \& Motion Disentanglement}
We evaluate how well CaSPR disentangles shape and motion as a result of the latent feature splitting $\z^C=[\z^C_{\text{ST}},\z^C_{\text{dyn}}]$. For this purpose, we transfer motion between two sequences by embedding both of them using TPointNet++, then taking the static feature $\z^C_{\text{ST}}$ from the first and the dynamic feature $\z^C_{\text{dyn}}$ from the second. \cref{fig:disentangle} shows qualitative results where each row is a different sequence; the first frame of the shape sequence is on the left, the point trajectories of the motion sequence in the middle, and the final CaSPR-sampled trajectories using the combined feature are on the right. If these features perfectly disentangle shape and motion, we should see the shape of the first sequence with the motion of the second after reconstruction. Apparently, the explicit feature split in CaSPR does disentangle static and dynamic properties of the object to a large extent.

\vspace{-9mm}\leavevmode\section{Conclusion}\vspace{-2mm}
\label{sec:conclude}
We introduced CaSPR, a method to canonicalize and obtain object-centric representions of raw point cloud sequences, which supports spatiotemporal sampling at arbitrary resolutions. We demonstrated CaSPR's utility on rigid and deformable object motion and in applications like spatiotemporal interpolation and estimating correspondences across time and instances. 

\parahead{Limitations and Future Work}
CaSPR leaves ample room for future exploration.
We currently only support batch processing, but online processing is important for real-time applications. Additionally, CaSPR is expensive to train.
Our canonicalization step requires dense supervision of T-NOCS labels which may not be available for real data.
While the network is well-suited for ST interpolation, the extrapolation abilities of CaSPR need further investigation. CaSPR is object-centric, and further work is needed to generalize to object collections and scenes. Additionally, outlier shapes can cause noisy sampling results and if the partial view of an object is ambiguous or the object is symmetric, TPointNet++ may predict a flipped or rotated canonical output.

Finally, using a single CNF for spatial sampling is fundamentally limited by an inability to model changes in topology~\cite{cornish2019relaxing,dupont2019augmented}. 
To capture fine-scale geometric details of shapes, this must be addressed.

\newpage
\section*{Broader Impact}
\label{sec:impact}
\vspace{-2mm}
CaSPR is a fundamental technology allowing the aggregation and propagation of dynamic point cloud information -- and as such it has broad applications in areas like autonomous driving, robotics, virtual/augmented reality and medical imaging.
We believe that our approach will have a mostly positive impact but we also identify potential undesired consequences below.

Our method will enhance the capabilities of existing sensors and allow us to build models of objects from sparse observations.
For instance, in autonomous driving or mixed reality, commonly used LIDAR/depth sensors are limited in terms of spatial and temporal resolution or sampling patterns.
Our method creates representations that overcome these limitations due to the capability to continuously sample in space and time.
This would enable these sensors to be cheaper and operate at lower spacetime resolutions saving energy and extending hardware lifespans.
Our approach could also be useful in spatiotemporal information propagation.
We can propagate sparse labels in the input over spacetime, leading to denser supervision.
This would save manual human labeling effort.

Like other learning-based methods, CaSPR can produce biased results missing the details in the input.
In a self driving scenario, if an input LIDAR point cloud only partially observes a pedestrian, CaSPR may learn representations that misses the pedestrian completely.
If real-world systems rely excessively on this incorrect representation it could lead to injuries or fatalities.
We look forward to conducting and fostering more research in other applications and negative impacts of our work.
\vspace{-2mm}

\section*{Acknowledgments and Funding Disclosure} 
This work was supported by grants from the Stanford-Ford Alliance, the SAIL-Toyota Center for AI Research, the Samsung GRO program, the AWS Machine Learning Awards Program, NSF grant IIS-1763268, and a Vannevar Bush Faculty Fellowship. The authors thank Michael Niemeyer for providing the code and shape models used to generate the warping cars dataset. Toyota Research Institute ("TRI") provided funds to assist the authors with their research but this article solely reflects the opinions and conclusions of its authors and not TRI or any other Toyota entity.

\bibliographystyle{splncs04}
\bibliography{references}

\clearpage
\newcounter{appendixsection}
\setcounter{appendixsection}{0}
\newcounter{appendixfigure}
\setcounter{appendixfigure}{0}
\newcounter{appendixtable}
\setcounter{appendixtable}{0}
\newcounter{appendixequation}
\setcounter{appendixequation}{0}

\renewcommand\thesection{\Alph{appendixsection}}
\renewcommand\thesubsection{\Alph{appendixsection}.\arabic{subsection}}
\renewcommand\thefigure{\Alph{appendixsection}\arabic{appendixfigure}}    
\renewcommand\thetable{\Alph{appendixsection}\arabic{appendixtable}}    
\renewcommand\theequation{\Alph{appendixsection}\arabic{appendixequation}}
\section*{Appendices}
We expand on discussions in~\cref{appendix:sec:discuss}, provide additional evaluations in~\cref{appendix:sec:addedexpts}, explain details of dataset generation and architecture implementation in ~\cref{appendix:sec:data} and ~\cref{appendix:sec:arch}, and give details of experiments from the main paper in~\cref{appendix:sec:exptdetails}.

\stepcounter{appendixsection}
\setcounter{appendixfigure}{0}
\setcounter{appendixtable}{0}
\setcounter{appendixequation}{0}
\section{Discussions}
\label{appendix:sec:discuss}

\noindent\textbf{Remarks on ODE-Nets } The requirements of homeomorphisms and differentiability impose certain limitations. First, neural ODEs lack a universal approximation capability as non-intersecting trajectories cannot learn to approximate arbitrary topologies~\cite{zhang2019approximation}\footnote{Augmented-Neural ODEs~\cite{runz2018maskfusion} propose to operate on a higher dimensional space as one workaround.}. On the other hand, it is also shown that this very property brings intrinsic robustness to ODE-Nets~\cite{yan2019robustness}. Moreover, the requirement of invertibility in CNFs is proven to hamper the approximation quality of the target distribution~\cite{cornish2019relaxing}. In fact, for a perfect recovery and likelihood evaluation, non-invertibility is a requirement~\cite{cornish2019relaxing}. Nonetheless, the extent to which these limitations restrict the applicability of Neural ODEs and CNFs is still an active research topic. 

\paragraph{Why can't we use existing point cloud networks as a canonicalizer?}
Extending PointNet++ to time (similar to MeteorNet~\cite{liu2019meteornet}) requires some form of a spatiotemporal neighborhood query or using time as an auxiliary input feature diminishing its contribution. Spatiotemporal neighborhood queries are undesirable as they necessitate difficult hyperparameter tuning and limit the network's ability to holistically understand the motion. For example, learning the arrow of time (as CaSPR does in Sec. 5 of the main paper) would be difficult when using local spatiotemporal queries. PointNet can somewhat remedy this by operating on the full 4D point cloud at once, treating time equally important as the spatial dimensions. However, we found that PointNet by itself is incapable of extracting descriptive local features, which are essential for an accurate mapping to T-NOCS.

\paragraph{On the arrow of time}  Due to the second law of thermodynamics, the entropy of an isolated system tends to increase with time, making the direction of time irreversible~\cite{lebowitz1993boltzmann} \ie~\emph{it is more common for a motion to cause multiple motions than for multiple motions to collapse into one consistent motion}~\cite{pickup2014seeing}. This causality is confirmed in computer vision by showing that the statistics of natural videos are not symmetric under time reversal~\cite{pickup2014seeing}. Any method processing spacetime inputs should then be sensitive to this direction so as to yield distinctive representations rather than being \emph{invariant} to it. As shown in the experiments of the main paper, thanks to the inclusion of timestamps and the Latent ODE advecting forward in time, CaSPR is highly aware of this unidirectionality and it is one of the reasons why it can extract robust spatiotemporal latent features.

\paragraph{On disentanglement} 
In the main paper, we have demonstrated experimentally that static and dynamic feature disentanglement is achieved to a large extent.
Note that CaSPR involves no mechanism that can guarantee a theoretically disentangled latent space such as the one of~\cite{zhao2019quaternion}. Our design \emph{softly} encourages the canonicalization network to respect the subspace nature by only advecting the dynamic feature with the ODE. Though this is not a CaSPR-specific drawback and many SoTA disentanglement networks rely upon the same intuition. 

\paragraph{Limitations}
Using a CNF to sample the object surface does come with some limitations as mentioned in prior work~\cite{yang2019pointflow} and discussed above. The inherent properties of CNFs may hamper the ability to capture fine-scale geometric detail. We observe this in chairs with back slats and other thin structures that are not captured by our Reconstruction CNF as shown in the left panel of~\cref{appendix:fig:failures}. Additionally, outlier shapes can cause noisy sampling results (shown in the middle). One current limitation of TPointNet++ is its inability to handle symmetry when canonicalizing a point cloud sequence. If the partial view of an object is ambiguous or the object is symmetric, TPointNet++ may predict a flipped or rotated canonical output as shown in the right panel.

\clearpage
\stepcounter{appendixfigure}
\begin{figure*}[t!]
    \centering
    \includegraphics[width=0.95\textwidth]{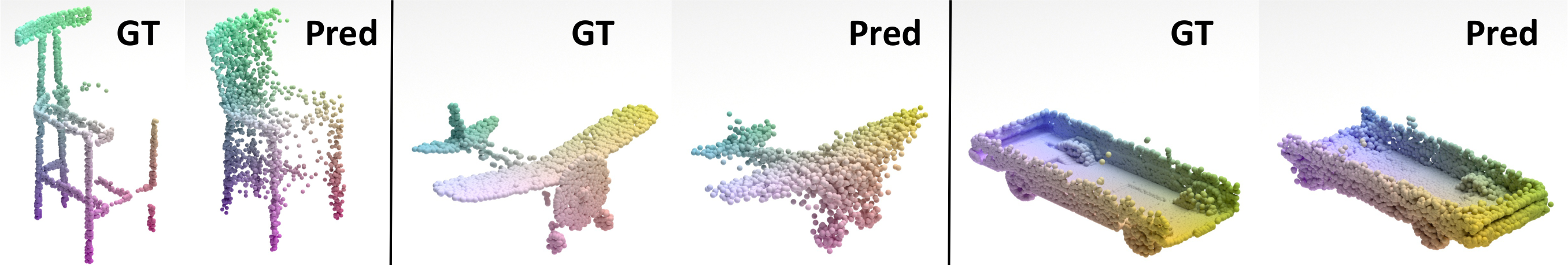}\vspace{-2mm}
    \caption{Failure cases of CaSPR. The CNF has difficulty capturing local details and very thin structures (left) along with uncommon shapes (middle). TPointNet++ has trouble with symmetry or ambiguity in partial views, resulting in reflected or rotated predictions (right).}
    \label{appendix:fig:failures}
\end{figure*}
\stepcounter{appendixsection}
\setcounter{appendixfigure}{0}
\setcounter{appendixtable}{0}
\setcounter{appendixequation}{0}
\stepcounter{appendixtable}
\begin{wraptable}{r}{0.5\textwidth}
\vspace{-50pt}
\caption{CaSPR ablations for reconstruction of rigid car sequences over 10 observed frames.}
\begin{center}
    \scalebox{0.9}{
    \begin{tabular}{ l c c c }
        \toprule
         \textbf{Method} & \textbf{CD} & \textbf{EMD} & \textbf{NFE} \\
        \midrule
        No  $\mathcal{L}_{c}$  &  0.605	& 11.482 & \textbf{38.2}  \\
        No Factorization  &  0.635 & 11.249 & 101.3  \\
        No Input Aug  & 0.577 & 10.253 & 38.9 \\
        Full Arch & \textbf{0.566} & \textbf{10.103}	& 39.6 \\
\bottomrule
\end{tabular}}
\end{center}
\vspace{-20pt}
\label{appendix:table:ablationresults}
\end{wraptable}
\section{Additional Evaluations}
\label{appendix:sec:addedexpts}

We provide evaluations of CaSPR omitted from the main paper due to space constraints. Please see Section 5 of the main paper for an explanation of evaluation metrics and the primary results.

\stepcounter{appendixtable}
\begin{wraptable}{r}{0.35\textwidth}
\vspace{-30pt}
\caption{Reconstruction errors with varying numbers of input points per frame for rigid car motion.}
\begin{center}
    \scalebox{0.9}{
    \begin{tabular}{ c c c }
        \toprule
         \textbf{Num Points} & \textbf{CD} & \textbf{EMD} \\
        \midrule
        2048  &  0.5657	& 10.1028  \\
        1024  &  \textbf{0.5486}	& \textbf{10.0339}  \\
        512  & 0.5904 & 10.5188  \\
        256 & 0.8222 & 13.8275 \\
        128 & 1.4730 & 20.7233 \\
\bottomrule
\end{tabular}}
\end{center}
\vspace{-5pt}
\label{appendix:table:sparsepoints}
\end{wraptable}

\subsection{Ablation Study}
\label{appendix:sec:ablationstudy}
We compare the full CaSPR architecture (\emph{Full Arch}) to multiple ablations in~\cref{appendix:table:ablationresults}. This includes: (i) not using the canonicalization loss (\emph{No $\mathcal{L}_c$}), (ii) not factorizing the latent ST feature and instead feeding the entire vector to the Latent ODE (\emph{No Factorization}), and (iii) using no pairwise terms (see~\cref{appendix:sec:arch}) as input augmentation (\emph{No Input Aug}). In addition to reconstruction metrics, we report the mean number of function evaluations (NFE) for the Latent ODE. This is the average number of times the ODE solver queries the dynamics network while integrating forward in time for a single sequence. Each method is trained on the rigid cars category and reconstructs all 10 input frames for evaluation. The full CaSPR architecture performs best. Note that the static/dynamic feature factorization is especially important to limit the complexity of Latent ODE dynamics. 

\subsection{Sparsity in Space and Time}
\label{appendix:sec:sparsityexpt}
We evaluate CaSPR's ability to reconstruct partial point cloud sequences from the rigid car category under sparsity in both space and time. Given 10 input frames, ~\cref{appendix:table:sparsepoints} shows the performance for reconstructing all 10 observed frames with a varying number of points available at each frame. 
\stepcounter{appendixtable}
\begin{wraptable}{r}{0.45\textwidth}
    \vspace{-15pt}
    \caption{Reconstruction errors with a varying number of observed input frames. }
    \centering
    \scalebox{0.725}{
    \begin{tabular}{ l c c c c }
            \toprule
        \multicolumn{1}{l}{} & \multicolumn{2}{c}{\emph{Observed}} & \multicolumn{2}{c}{\emph{Unobserved}} \\
        \textbf{Num Observed} &  \textbf{CD} & \multicolumn{1}{c}{ \textbf{EMD}} &  \textbf{CD} &  \textbf{EMD} \\
        \midrule
        10 steps & \textbf{0.5657} & \textbf{10.1028}	& --- & --- \\
        7 steps & 0.5701 & 10.3406 & \textbf{0.5609} & \textbf{10.0304} \\
        5 steps & 0.5664 & 10.4310 & 0.5620 & 10.3374 \\
        3 steps & 0.5904 & 11.4641 & 0.5837 & 11.2586 \\
        2 steps & 0.7095 & 14.8348 & 0.7233 & 16.3499 \\
        \bottomrule
    \end{tabular}
    }
    \vspace{-7pt}
\label{appendix:table:sparsetime}
\end{wraptable}
Performance is consistent until 256 or fewer points are given at which point it drops off rapidly. ~\cref{appendix:table:sparsetime} shows performance when varying the number of available observed timesteps for each test sequence. Observed timesteps are distributed as evenly as possible over the 10-step sequence for this evaluation. Performance is stable even with 3 observed frames, but does significantly drop when only 2 frames are given (\ie~the first and last steps). 

\stepcounter{appendixtable}
\begin{wraptable}{r}{0.35\textwidth}
\vspace{-47pt}
\caption{Reconstruction errors for longer sequences on rigid car data.}
\begin{center}
    \scalebox{0.8}{
    \begin{tabular}{ c c c }
        \toprule
         \textbf{Test Seq Length} & \textbf{CD} & \textbf{EMD} \\
        \midrule
        10  &  0.566 & \textbf{10.103} \\
        25  & \textbf{0.534}	& 10.815  \\
\bottomrule
\end{tabular}}
\end{center}
\vspace{-10pt}
\label{appendix:table:longseq}
\end{wraptable}

\subsection{Reconstructing Longer Sequences}
We evaluate CaSPR when trained on the rigid motion car dataset with sequences of 25 frames (rather than 10 as in the main paper). During training, we randomly subsample 10 frames (rather than 5) from each sequence, and evaluate with the full 25-frame sequence as input (rather than 10). ~\cref{appendix:table:longseq} shows reconstruction performance compared to the model in the main paper which uses the 10-frame sequence dataset. We see there is a minimal difference in performance, indicating CaSPR is capable of handling longer-horizon motion.

\stepcounter{appendixtable}
\begin{wraptable}{r}{0.4\textwidth}
\vspace{-45pt}
\caption{Reconstruction errors training on all categories jointly.}
\begin{center}
    \scalebox{0.8}{
    \begin{tabular}{ c c c c }
        \toprule
         \textbf{Train Data} & \textbf{Test Data} & \textbf{CD} & \textbf{EMD} \\
        \midrule
        Cars  &  Cars & \textbf{0.566} & \textbf{10.103}   \\
        All  &  Cars & 0.728 & 13.631   \\
        \midrule
        Chairs  &  Chairs &  \textbf{0.715} & \textbf{13.009}  \\
        All  &  Chairs & 1.231 & 15.632  \\
        \midrule
        Airplanes  &  Airplanes &  \textbf{0.231} & \textbf{6.026}   \\
        All  &  Airplanes &  0.391 & 8.213  \\
        \midrule
        All & All & 0.798 & 12.578 \\
\bottomrule
\end{tabular}}
\end{center}
\vspace{-15pt}
\label{appendix:table:allcats}
\end{wraptable}

\subsection{Multi-Category Model}
We evaluate CaSPR when trained on all shape categories together: cars, chairs, and airplanes. This determines the extent of the category-level restriction on our method. Results compared to models trained on each category separately are shown in~\cref{appendix:table:allcats}. Models are evaluated by reconstructing all 10 observed time steps. As expected, there is a performance drop when training a single joint model, however errors are still reasonable and in most cases better than the \emph{PointFlow} baseline in terms of EMD (see Tab. 2 in main paper).

\stepcounter{appendixtable}
\begin{wraptable}{r}{0.35\textwidth}
\vspace{-60pt}
\caption{Canonicalization performance for deforming cars.}
\vspace{-3mm}
\begin{center}
    \scalebox{0.8}{
    \begin{tabular}{ c c c }
        \toprule
         \textbf{Method} & \textbf{Spatial Err} & \textbf{Time Err} \\
        \midrule
        Identity  &  0.0583	& \textbf{0.0000} \\
        TPointNet++  &  \textbf{0.0221} & 0.0012  \\
\bottomrule
\end{tabular}}
\end{center}
\vspace{-10pt}
\label{appendix:table:canondeform}
\end{wraptable}

\subsection{Canonicalizing for Deformation}
We evaluate the ability of TPointNet++ to canonicalize non-rigid transformations. Given a deforming car sequence from the Warping Cars dataset, the task is to remove the deformation at each step, leaving the base shape without any warping. To achieve this, we train TPointNet++ with $\mathcal{L}_{c}$ only, and supervise every step in a sequence with the same GT canonical point cloud that contains no deformation. Note that Warping Cars is already canonical in terms of rigid transformations, so the network needs to learn to factor out non-rigid deformation only. 
Results are shown in~\cref{appendix:table:canondeform} where we compare TPointNet++ to a baseline that simply copies the input points to the output (\emph{Identity}, which performs reasonably since there is no rigid transformation). \emph{Identity} trivially gives a perfect time error, but TPointNet++ achieves a much lower spatial error, effectively removing the deformation from each step. This is qualitatively shown in~\cref{appendix:fig:deformresults}. This strategy of canonicalization offers an explicit way to extract temporal correspondences over time, rather than relying on the CNF to naturally exhibit correspondences (main paper Sec. 5).

\subsection{Label Propagation through Canonicalization}
We evaluate the ability of T-NOCS canonicalization to establish correspondences by propagating point-wise labels both throughout a sequence and to new sequences of different object instances. Given a semantic segmentation of the partial point cloud at the \emph{first} frame of a sequence at time $s_1$, the first task is to label all subsequent steps in the sequence at times $s_2, \dots, s_k$, \ie~propagate the segmentation forward in time. Secondly, we want to label all frames of sequences containing \emph{different} object instances \ie~propagate the segmentation to different objects of the same class. We achieve both through canonicalization with TPointNet++: all frames in each sequence are mapped to T-NOCS, then unknown points are labeled by finding the closest point in the given labeled frame at $s_1$. If the closest point in $s_1$ is not within a distance of $0.05$ in the canonical space, it is marked ``Unknown". This may happen if part of the shape is not visible in the first frame due to self-occlusions.

\stepcounter{appendixfigure}
\begin{figure*}[t!]
    \centering
    \includegraphics[width=\textwidth]{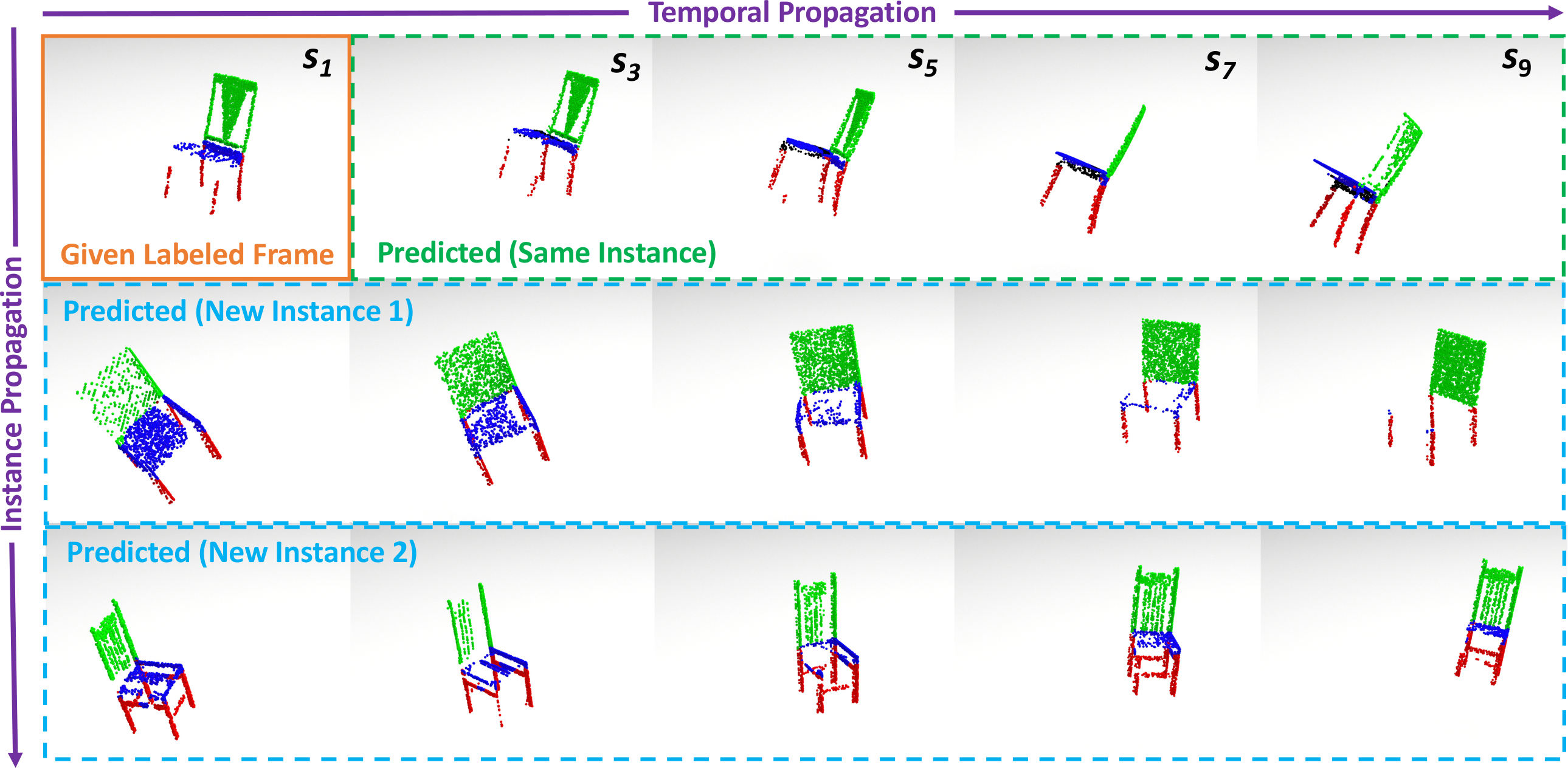}\vspace{-2mm}
    \caption{Example of semantic segmentation label propagation over time and across instances through T-NOCS canonicalization. The given labels in the first frame of the top sequence (orange box) are transferred to later frames in the same sequence (green dashed box) and to other sequences with different object instances (blue dashed boxes) by comparing to the labeled frame in the shared canonical space.}
    \label{appendix:fig:labelprop}
    \vspace{-\baselineskip}
\end{figure*}

\stepcounter{appendixtable}
\begin{wraptable}{r}{0.45\textwidth}
\vspace{-20pt}
\caption{Segmentation label propagation performance. \emph{Total Acc} is point-wise accuracy over all points; \emph{Known Acc} is only for points that our method successfully labels.}
\vspace{-10pt}
\begin{center}
    \scalebox{0.8}{
    \begin{tabular}{ c c c c }
        \toprule
        \textbf{Task} &  \textbf{Category} & \textbf{Total Acc} & \textbf{Known Acc} \\
        \midrule
        Temporal & Chairs  &  0.9419 & 0.9804 \\
        Propagation & Airplanes  &  0.9580 & 0.9676  \\
        \midrule
        Instance & Chairs  &  0.6553 & 0.8425 \\
        Propagation & Airplanes  &  0.7744 & 0.8006  \\
\bottomrule
\end{tabular}}
\end{center}
\vspace{-20pt}
\label{appendix:table:labelprop}
\end{wraptable}

Results of this label propagation for a subset of the chairs (1315 sequences) and airplanes (1215 sequences) categories of the rigid motion test set are shown in ~\cref{appendix:table:labelprop}. We report median point-wise accuracy over all points (\emph{Total Acc}) and for points successfully labeled by our approach (\emph{Known Acc}). For the instance propagation task, we randomly use $1/3$ of test sequences as ``source" sequences where the first frame is labeled, and the other $2/3$ are ``target" sequences to which labels are propagated. In this case, accuracy is reported only for target sequences. Qualitative results are shown in~\cref{appendix:fig:labelprop}.

\subsection{Extrapolating Motion}
We evaluate CaSPR's ability to extrapolate future motion without being explicitly trained to do so. In particular, the model is given the first 5 frames in each sequence and must predict the following 5 frames. The ability to predict future motion based on the learned prior would be valuable in real-time settings. We evaluate the already-trained full reconstruction models for each object category (from Tab. 2 in the main paper). \stepcounter{appendixtable}
\begin{wraptable}{r}{0.35\textwidth}
\vspace{-15pt}
\caption{Reconstruction of extrapolated frames.}
\vspace{3pt}
\centering
\scalebox{0.7}{
\begin{tabular}{ c c c c c }
        \toprule
    & \multicolumn{2}{c}{\emph{5 Observed}} & \multicolumn{2}{c}{\emph{5 Extrapolated}}  \\
    \textbf{Category} &  \textbf{CD} & \multicolumn{1}{c}{ \textbf{EMD}} &  \textbf{CD} &  \textbf{EMD} \\
    \midrule
    Cars & 0.597 & 9.833 & 1.023 & 21.055 \\
    Chairs & 0.687 & 12.502 & 1.010 & 20.648 \\
    Airplanes & 0.224 & 5.719 & 0.286 & 9.625 \\
    \bottomrule
\end{tabular}
}
\vspace{-15pt}
\label{appendix:table:extrapolation}
\end{wraptable} Note that these models are supervised with observed frames - they are not trained to predict unseen future states. Results are shown in~\cref{appendix:table:extrapolation}. Clearly there is a sharp performance drop between observed and extrapolated frames as we might expect, though performance is actually on par with the AtlasNet baseline (Tab. 2, main paper) in some cases. We note that qualitatively, the model produces reasonable future motion based on what it has seen and even hallucinates unseen parts of the shape, though it cannot handle sudden changes in direction.
\stepcounter{appendixsection}
\setcounter{appendixfigure}{0}
\setcounter{appendixtable}{0}
\setcounter{appendixequation}{0}
\section{Datasets Details}
\label{appendix:sec:data}

\paragraph{Rigid Motion Dataset} Please see Section 5 of the main paper for an introduction to our new dataset containing rigid motion for ShapeNet~\cite{chang2015shapenet} cars, chairs, and airplanes. This simulated dataset gives us the ability to capture a wide range of trajectories and acquire the necessary inputs and supervision to train and evaluate CaSPR. 

\stepcounter{appendixfigure}
\begin{wrapfigure}{r}{0.2\textwidth}
    \vspace{-6mm}
\includegraphics[width=0.2\textwidth]{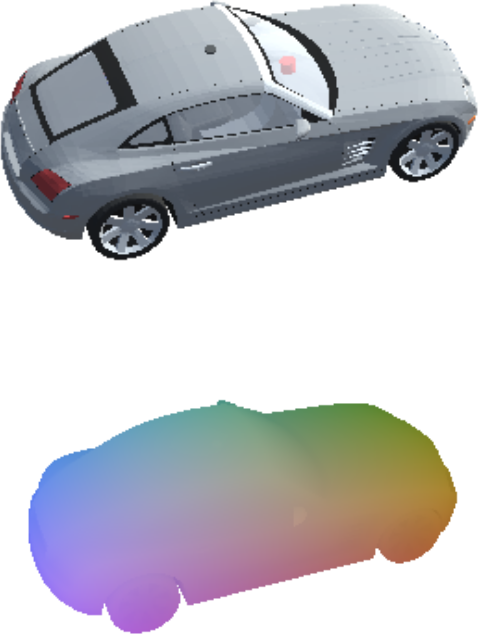}\vspace{-4mm}
    \caption{NOCS map from rigid motion dataset.}
    \label{appendix:fig:nocsex}
    \vspace{-10mm}
\end{wrapfigure}

We generate these motions within the Unity game engine\footnote{https://unity.com/}. For each object instance, we simulate a camera trajectory around the object (placed at the origin) that starts at a random location and continues for 50 timesteps. The camera always points towards the origin and its location is parameterized as a point on the surface of a sphere centered at the origin: by a longitudinal and latitudinal angle along with a radius. To produce a trajectory, each of these parameters is gradually increased or decreased independently. When a parameter reaches a set limit, its direction is reversed, producing interesting and challenging motions. At each step of the trajectory, a depth map and NOCS map~\cite{wang2019normalized} are rendered from the current camera view. An example NOCS map from the dataset is shown in ~\cref{appendix:fig:nocsex}. Example camera trajectories and the resulting aggregate canonical point cloud are shown in~\cref{appendix:fig:campose}.

The rendered frames are further processed to produce the final dataset of raw depth and canonical T-NOCS point cloud sequences. The rendered trajectory for each object instance is split into 5 sequences (with 10 steps each). 4096 pixels on the object are uniformly sampled from each depth map to extract \emph{raw} partial point cloud sequences in the world (camera) frame that are used as the input to CaSPR. Examples of these partial sequences are shown in~\cref{appendix:fig:dataex}. Each input point cloud in a sequence is given a timestamp in uniform steps from 0.0 to 5.0. The same sampled pixels are taken from the NOCS map to extract a corresponding canonical partial point cloud and given a timestamp from 0.0 to 1.0: this represents the supervision for CaSPR. In total, the car category contains $2527$ object instances ($12,635$ sequences), chairs contains $5000$ objects ($25,000$ sequences), and airplanes has $4045$ objects ($20,225$ sequences). Each category is split 80/10/10 into train/val/test. The val/test sets are entirely made up of object instances and camera motions that do not appear in the training split. 

\stepcounter{appendixfigure}
\begin{figure*}[t!]
    \centering
    \includegraphics[width=0.9\textwidth]{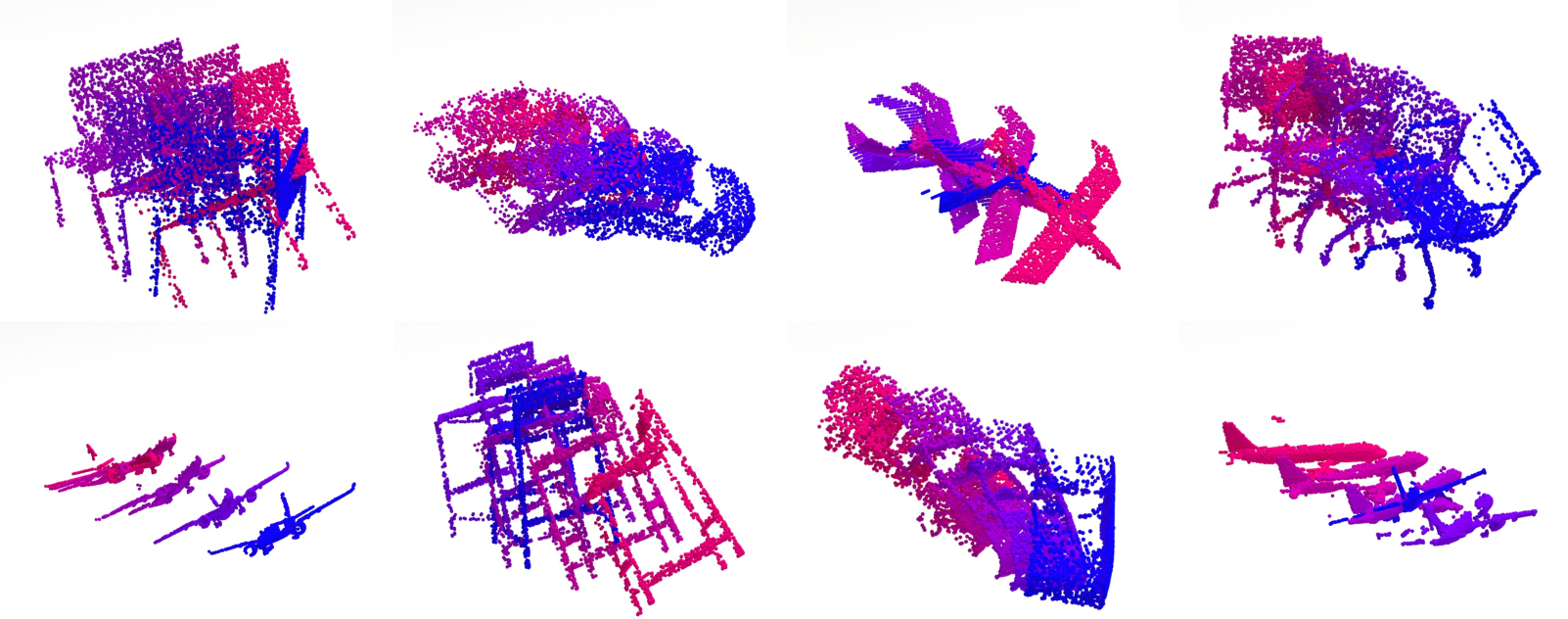}\vspace{-2mm}
    \caption{Examples from the rigid motion dataset. Partial point cloud sequences resulting from rendered data depth maps are shown; color shifts from blue to red over time.}
    \label{appendix:fig:dataex}
    \vspace{-\baselineskip}
\end{figure*}

Note that during training and inference, only a subset of the available 4096 points at each step in the dataset are used, as detailed in the main paper (usually 1024 during training and 2048 during evaluation). Additionally, during training a subset of the available 10 frames are randomly sampled from each sequence, giving non-uniform step sizes between observations. These subsampled sequences are shifted so that $s_1 = 0.0$ before being given to CaSPR, making things practically easier as it ensures that the Latent ODE always starts from $\overbar{t}_1 = 0$ for any sequence in a batch.

\paragraph{Warping Cars Dataset} In Section 5.1 of the main paper (``Non-Rigid   Reconstruction and Temporal Correspondences"), we use a variation of the Warping Cars dataset from Occupancy Flow (OFlow)~\cite{niemeyer2019occupancy}. We generate our version of this dataset with code kindly provided by the authors of that work. The dataset contains the same car models as our rigid motion dataset, however they are watertight versions that allow determining occupancy, which is needed to train OFlow. Same as the rigid motion dataset, we generate 5 sequences for each car instance with 10 frames of motion each. Consistent with the OFlow paper, we sample 100k points per sequence on the surface of the object that are in correspondence over time and can be used as inputs to CaSPR and OFlow; we also sample 100k points in the unit cube containing the object with corresponding occupancy labels for OFlow. Note that this data gives point clouds on the \emph{complete} object rather than the partial surface, and there is no rigid motion in the dataset -- only deformation. This means the sequences are already canonical in the sense that cars are consistently aligned and scaled. We also use input timestamps from 0.0 to 1.0, so the data is already canonical in time as well.

\stepcounter{appendixsection}
\setcounter{appendixfigure}{0}
\setcounter{appendixtable}{0}
\setcounter{appendixequation}{0}
\section{Implementation Details}
\label{appendix:sec:arch}
We next cover additional architectural and training details of our method. Please see Section 4.1 of the main paper for the primary discussion of our architecture and training procedure. We implement our method using PyTorch\footnote{https://pytorch.org/}.

\paragraph{TPointNet++}
The PointNet~\cite{qi2017pointnet} component operates on the entire 4D input point cloud and extracts a 1024-dimensional global feature and 64-dimensional per-point features. We use the vanilla classification PointNet architecture with 3 shared fully-connected (FC) layers (64, 128, 1024), ReLU non-linearities, and a final max-pool function. The per-point features come from the output of the first FC layer, while the global feature is the output of the max-pool. We do not use the input or feature transform layers, and replace all batch normalization with group normalization~\cite{wu2018group} using 16 groups, which is crucial to good performance with small batch sizes.

The PointNet++~\cite{qi2017pointnet++} component operates on each frame of the point cloud sequence independently and does not receive the timestamp as input. The input points to this part of the network are augmented with pairwise terms $x^2$, $y^2$, $z^2$, $xy$, $yz$, and $xz$, which we found improves reconstruction performance (see~\cref{appendix:sec:ablationstudy}). We use a modified version of the segmentation architecture which contains 5 set abstraction (SA) layers (PointNet dimensions, radii, number points out): $([[16, 16, 32],[32, 32, 64]], [0.8, 0.4], 1024) \rightarrow ([[32, 32, 64],[32, 32, 64]], [0.4, 0.2], 512) \rightarrow ([[64, 64, 128],[64, 96, 128]], [0.2, 0.1], 256) \rightarrow ([[128, 256, 256],[128, 256, 256]], [0.1, 0.05], 64) \rightarrow ([[256, 256, 512],[256, 256, 512]], [0.05, 0.02], 16)$. These are followed by 5 feature propagation (FP) layers which each have 2 layers with hidden size $512$, and a final shared MLP with layers $(512, 512)$ to produce the final per-point 512-dimensional local feature. ReLU non-linearities are used throughout, and we again replace all batch normalization with group normalization~\cite{wu2018group} using 16 groups. 

The final shared MLP which processes the concatenated features from PointNet and PointNet++ also uses group normalization and ReLU. 

There are a few things of note with this architecture. First of all, it avoids any spatiotemporal neighborhood queries since time is handled entirely with PointNet which treats the timestamps as an additional spatial dimension. This allows the network to decide which time windows are most important to focus on. Second, the architecture can easily generalize to sequences with differing numbers of points and frames since both are processed almost entirely independently (the only components affected by changing these at test-time are the PointNet max-pooling and the PointNet++ spatial neighborhood queries). 

\paragraph{Latent ODE}
The Latent ODE is given a 64-dimensional latent state $\z_0\triangleq\z^C_{\text{dyn}}$ which can be advected to any canonical timestamp from 0.0 to 1.0. The dynamics of the Latent ODE is an MLP with 3 hidden layers (512, 512, 512) which uses Tanh non-linearities. We use the \emph{torchdiffeq} package\footnote{https://github.com/rtqichen/torchdiffeq}~\cite{chen2018neural} which implements both the ODE solver along with the adjoint method to enable backpropagation. We use the \emph{dopri15} solver which is an adaptive-step Runge-Kutta 4(5) method. We use a relative tolerance of 1e-3 and absolute tolerance of 1e-4 both at training and test time.

\paragraph{Reconstruction CNF}
Our reconstruction CNF adapts the implementation of FFJORD~\cite{grathwohl2019ffjord} for point clouds from PointFlow~\cite{yang2019pointflow}. The dynamics of the CNF are parameterized by a neural network that uses 3 hidden \emph{ConcatSquashLinear} layers (512, 512, 512), which are preceeded and followed by a \emph{Moving Batch Normalization} layer. We use Softplus non-linearities after each layer. Please see ~\cite{yang2019pointflow} for full details. In short, each layer takes as input the current hidden state (512-dimensional at hidden layers or 3-dimensional $x,y,z$ at the first layer), the conditioning shape feature (1600-dimensional in CaSPR), and the current time of the flow (scalar), and uses this information to update the hidden state (or output the 3-dimensional derivative at the last layer). The ODE is again solved using \emph{dopri15}, this time with both a relative and absolute tolerence of 1e-5. We use the adjoint method for backpropagation and jointly optimize for the final flow time $T$ along with the parameters of network.

\paragraph{Training and Inference}
In practice, the full loss function is $\mathcal{L}= w_r \mathcal{L}_{r} + w_c \mathcal{L}_{c}$ where the contributions of the reconstruction and canonicalization terms are weighted as $w_r = 0.01$ and $w_c = 100$ as to be similar scales. No weight decay is used. We use the Adam~\cite{kingma2014adam} optimizer ($\beta_1 = 0.9$, $\beta_2=0.999$) with a learning rate of 1e-4. During training, we periodically compute the validation set loss, and after convergence use the weights with the best validation performance as the final trained model. The number of epochs trained depends on the dataset and the task. We train across up to 4 NVIDIA Tesla V100 GPUs which allows for a batch size of up to 20 sequences of 5 frames each. As noted in previous work~\cite{yang2019pointflow}, solving and backpropagating through ODEs (two in our case: Latent and CNF) results in slow training: it takes about 5 days for the full CaSPR architecture using the multi-gpu setup. The full CaSPR network contains about 16 million trainable parameters. Inference for a 10-step sequence of rigid car motion with 2048 points at each step takes on average 0.598 seconds.
\stepcounter{appendixsection}
\setcounter{appendixfigure}{0}
\setcounter{appendixtable}{0}
\setcounter{appendixequation}{0}
\section{Experimental Details and Supplemental Results}
\label{appendix:sec:exptdetails}
Here we give details of experiments shown in Section 5 of the main paper along with some supporting results for these experiments (\eg~means, standard deviations, and visualizations).

\paragraph{Evaluation Procedure}
To evaluate reconstruction error, we use the Chamfer Distance (CD) and Earth Mover's Distance (EMD). For our purposes, we define the CD and EMD between two point clouds $\Xset_1, \Xset_2$ each with $N$ points as
$$d_{C D}\left(\Xset_{1}, \Xset_{2}\right)= \dfrac{1}{N} \sum_{\mathbf{x}_1 \in \Xset_{1}} \min _{\mathbf{x}_2 \in \Xset_{2}}\|\mathbf{x}_1-\mathbf{x}_2\|_{2}^{2}+ \dfrac{1}{N}\sum_{\mathbf{x}_2 \in \Xset_{2}} \min _{\mathbf{x}_1 \in \Xset_{1}}\|\mathbf{x}_1-\mathbf{x}_2\|_{2}^{2}$$
$$d_{EMD}\left(\Xset_{1}, \Xset_{2}\right)=\min _{\phi: \Xset_{1} \rightarrow \Xset_{2}} \dfrac{1}{N} \sum_{\mathbf{x}_1 \in \Xset_{1}}\|\mathbf{x}_1-\phi(\mathbf{x}_1)\|_{2}^{2}$$

where $\phi: \Xset_{1} \rightarrow \Xset_{2}$ is a bijection. In practice, we use a fast approximation of the EMD based on~\cite{bertsekas1985emd}. Both CD and EMD are always reported multiplied by $10^3$.

\stepcounter{appendixtable}
\begin{wraptable}[12]{r}{0.5\textwidth}
\vspace{-20pt}
\caption{\small Canonicalization performance mean and (standard deviation). Supplements Tab. 1 in the main paper.}
\vspace{-8pt}
\begin{center}
\scalebox{0.7}{
\setlength{\tabcolsep}{4pt}
\begin{tabular}{ r c c c }
\toprule
 \textbf{Method} & \textbf{Category} & \textbf{Spatial Err} & \textbf{Time Err} \\
\midrule
MeteorNet &  Cars & 0.0834 (0.0801) & \textbf{0.0002} (0.0015) \\
PointNet++ No Time &   &  0.0649 (0.0468)	& --- \\
PointNet++ w/ Time &   &  0.0715 (0.0811) & 0.0006 (0.0012)  \\
PointNet &   & 0.0485 (0.0952) & 0.0016 (0.0015) \\
TPointNet++ No Aug &  &  0.0225 (0.0501) & 0.0015 (0.0014) \\
TPointNet++ No Time &   &  \textbf{0.0224} (0.0570) & --- \\
\midrule
TPointNet++ &  Cars &  0.0229 (0.0617) & 0.0013 (0.0012)  \\
TPointNet++ &  Chairs & 0.0162 (0.0337) & 0.0008 (0.0006) \\
TPointNet++ &  Airplanes &  0.0148 (0.0412) & 0.0009 (0.0007)  \\
\bottomrule
\end{tabular}}
\end{center}
\label{appendix:table:canonmeans}
\end{wraptable}
 As noted in the main paper, for these reconstruction metrics and the canonicalization error metrics, we report the median values over all test frames. This is motivated by the fact that ShapeNet~\cite{chang2015shapenet} contains some outlier shapes which result in large errors that unfairly bias the mean and do not accurately reflect comprehensive method performance. For completeness, we also report mean and standard deviation for these metrics in this document for main paper experiments. Note that CD and EMD, along with the spatial canonicalization error, are all reported in the canonical space where the shape lies within a unit cube. This helps intuit the severity of reported errors. 
 
 Although we randomly subsample 1024 points at each frame for training, during evaluation we always use the same 2048 points (unless specifically stated otherwise) to make evaluation consistent across compared methods. Unless otherwise stated, CaSPR and all compared baselines reconstruct the same number of points as in the input (\eg~for evaluation, each input frame has 2048 points, so we sample 2048 points from our Reconstruction CNF). 
 
\paragraph{Canonicalization}
In this experiment, we train TPointNet++ by itself with only the canonicalization loss $\mathcal{L}_{c}$ on each category of the rigid motion dataset. In order to make the number of parameters comparable across all baselines, we use hidden layers of size 1024 (rather than 1600) in the final shared MLP for the full TPointNet++ architecture only. We compare to the following baselines which are all trained with the same $\mathcal{L}_{c}$:\vspace{-2mm}
\begin{itemize}[leftmargin=*]
    \item \emph{MeteorNet}~\cite{liu2019meteornet}: A recent method that extends PointNet++ to process point cloud sequences through spatiotemporal neighborhood queries. We adapt the \textit{MeteorNet-seg} version of the architecture with \textit{direct grouping} for our task by adding an additional \textit{meteor direct module} layer, as well as two fully connected layers before the output layer. Additionally, we slightly modify feature sizes to make the model capacity comparable to other methods. We found the spatiotemporal radii hyperparameters difficult to tune and in the end we opted for 10 uniformly sampled radii between $(0.03,0.05)$ in the first layer, which were doubled in each subsequent layer. 
    \item \emph{PointNet++ No Time}: An ablation of TPointNet++ that removes the PointNet component. This leaves PointNet++ processing each frame independently followed by the shared MLP, and therefore has no notion of time.
    \item \emph{PointNet++ w/ Time}: This is the same ablation as above, but modified so that the PointNet++ receives the timestamp of each point as an additional input feature. Note that local neighborhood queries are still performed only on spatial points, but they may be across timesteps so we use increased radii of $(0.05, 0.1, 0.2, 0.6, 1.2, 2.0)$.  This baseline represents a naive way to incorporate time, but dilutes its contributions since it is only an auxiliary feature. 
    \item \emph{PointNet}: An ablation of TPointNet++ that removes the PointNet++ component. This leaves only PointNet operating on the full 4D spatiotemporal point cloud. This baseline treats time equally as the spatial dimensions, but inherently lacks local geometric features. 
    \item \emph{TPointNet++ No Time}: An ablation of TPointNet++ that only regresses the spatial part of the T-NOCS coordinate (and not the normalized timestamp). This baseline still takes the timestamps as input, it just doesn't regress the last time coordinate. 
    \item \emph{TPointNet++ No Aug}: An ablation of TPointNet++ that does not augment the input points to PointNet++ with pairwise terms as described previously. This baseline was omitted from the main paper for brevity, so a comparison of median performance is shown in~\cref{appendix:table:canonresults_noaug}.
\end{itemize}

\stepcounter{appendixtable}
\begin{wraptable}{r}{0.45\textwidth}
\vspace{-15pt}
\caption{\small Canonicalization performance without input augmentation.}
\begin{center}
\scalebox{0.7}{
\begin{tabular}{ r c c c }
\toprule
 \textbf{Method} & \textbf{Category} & \textbf{Spatial Err} & \textbf{Time Err} \\
\midrule
No Aug &  Cars &  0.0138 & 0.0012 \\
Full Arch &  Cars &  \textbf{0.0118}	& \textbf{0.0011}  \\
\bottomrule
\end{tabular}}
\end{center}
\vspace{-5pt}
\label{appendix:table:canonresults_noaug}
\end{wraptable}

Each model is trained for 220 epochs on the cars category. TPointNet++ is trained for 120 and 70 epochs on the airplanes and chairs categories, respectively, due to the increased number of objects. Median canonicalization errors are in Tab. 1 of the main paper; the mean and standard deviations are shown in~\cref{appendix:table:canonmeans}.

\paragraph{Representation and Reconstruction}
In this experiment, we compare the full CaSPR architecture to two baselines on the task of reconstructing a partial point cloud sequence.

The baselines represent one alternative to achieve spatial continuity, and one to achieve temporal continuity. The \emph{CaSPR-Atlas} baseline is the full CaSPR architecture as described, but replaces the Reconstruction CNF with an AtlasNet~\cite{groueix2018atlasNet} decoder. We use the same decoder as the original AtlasNet. This decoder contains 64 MLPs, each responsible for transforming a patch to the partial visible surface at a desired timestep. Each MLP contains 4 hidden layers $(1600,1600,800,400)$ with Tanh activation functions. This version of CaSPR is still trained with the auxiliary canonicalization task ($\mathcal{L}_{c}$ loss), but the reconstruction loss is now a Chamfer distance since AtlasNet does not support likelihood evaluations like a CNF. We use group normalization~\cite{wu2018group} instead of batch normalization within the decoder to improve performance with small batch sizes.

\stepcounter{appendixtable}
\begin{table}[t!]
    \caption{Partial surface sequence reconstruction results showing mean and (standard deviation). Supplements Tab. 2 in the main paper.}
    \centering
    \scalebox{0.725}{
    \begin{tabular}{ l c  c c | c c c c }
            \toprule
        & \multicolumn{1}{l}{} & \multicolumn{2}{c}{\emph{10 Observed}} & \multicolumn{2}{c}{\emph{3 Observed}} & \multicolumn{2}{c}{\emph{7 Unobserved}} \\
        \textbf{Method} &  \textbf{Category} &  \textbf{CD} & \multicolumn{1}{c}{ \textbf{EMD}} &  \textbf{CD} &  \textbf{EMD} &  \textbf{CD} &  \textbf{EMD} \\
        \midrule
        PointFlow &  Cars &  \textbf{0.537} (0.272) & 15.986 (11.130) &  \textbf{0.538} (0.270) &   \textbf{15.967} (11.065) &  \textbf{0.700} (0.732) & 17.362 (12.276) \\
        CaSPR-Atlas & Cars &  0.814	(1.729) & 26.922 (28.562) &	0.874 (2.051) & 29.171 (29.479) & 0.853 (1.705) & 26.416 (27.582) \\
        CaSPR &  Cars & 0.795 (1.048) &  \textbf{14.242} (21.619) & 0.846 (1.261)	& 16.564 (24.296) & 0.824 (1.108) &  \textbf{16.217} (23.011) \\
        \midrule
        PointFlow &  Chairs &  \textbf{0.907} (0.519) & 20.254 (11.938) & \textbf{0.907} (0.514) & 20.225 (11.899) & 1.245 (1.299)	& 21.971 (13.417) \\
        CaSPR-Atlas & Chairs &  1.007 (1.243) & 54.406 (24.970) & 1.030 (1.221) & 54.827 (25.250) & 1.061 (1.277)	& 52.964 (24.355) \\
        CaSPR &  Chairs & 1.013 (1.426) & \textbf{15.287} (9.837) & 0.972 (1.498) & \textbf{15.757} (11.154) & \textbf{1.000} (1.542) & \textbf{16.145} (11.620) \\
        \midrule
        PointFlow &  Airplanes & \textbf{0.367} (0.366) & 11.852 (8.768) & \textbf{0.366} (0.363) & 11.862 (8.725) & \textbf{0.446} (0.527) & 12.335 (9.146)  \\
        CaSPR-Atlas & Airplanes &  0.587 (1.196) & 23.444 (17.386) & 0.653 (1.369) & 23.165 (16.932) & 0.663 (1.400) & 22.661 (16.853) \\
        CaSPR &  Airplanes &  0.536 (1.468) & \textbf{8.827} (12.650) & 0.536 (1.682) & \textbf{8.992} (13.219) & 0.530 (1.673) & \textbf{9.031} (12.792) \\
        \bottomrule
    \end{tabular}
    }
\label{appendix:table:reconmeans}
\end{table}
\stepcounter{appendixfigure}
\begin{figure*}[t!]
    \centering
    \includegraphics[width=\textwidth]{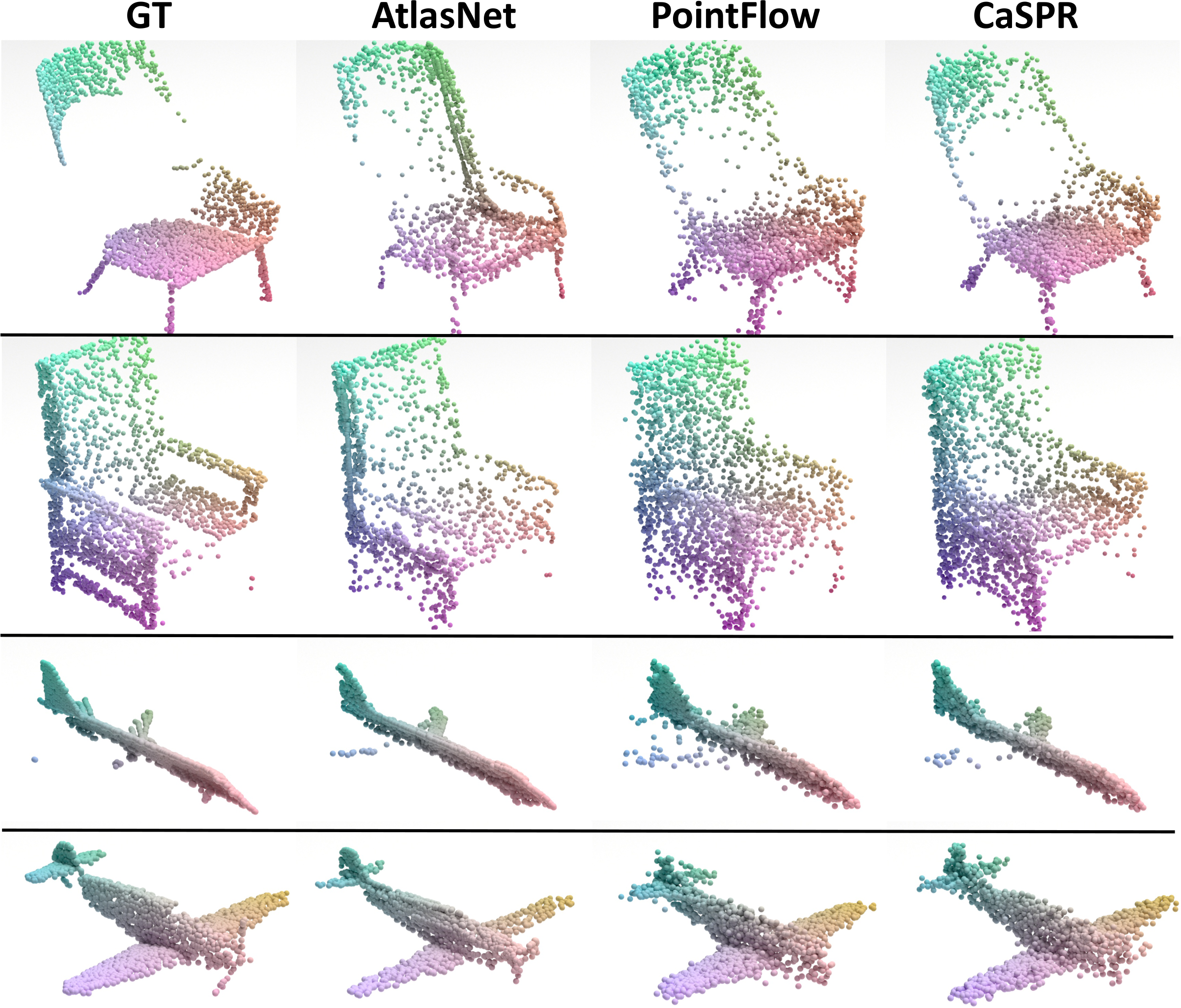}\vspace{-2mm}
    \caption{Reconstruction performance of the \emph{CaSPR-Atlas} and \emph{PointFlow} baselines compared to the full CaSPR model. Each row shows a frame from a different 10-step rigid motion sequence.}
    \label{appendix:fig:pointflowatlascompare}
    \vspace{-\baselineskip}
\end{figure*}

The \emph{PointFlow}~\cite{yang2019pointflow} baseline uses their deterministic autoencoder architecture. This follows the autoencoding evaluations from the original paper and uses a PointNet-like encoder to extract a shape feature, which conditions a CNF decoder. This version of the model is trained only with the reconstruction likelihood objective from the CNF, and does not use the various losses associated with the VAE formulation of their architecture. To make it a fair comparison, we increase the size of the shape feature bottleneck to 1600. The CNF decoder uses a dynamics MLP with 3 hidden layers of size $(512, 512, 512)$, just like CaSPR. Also like CaSPR, we train \emph{PointFlow} with a learning rate of 1e-4, which we found to decrease the complexity of dynamics and therefore training time.

\stepcounter{appendixfigure}
\begin{figure*}[t!]
    \centering
    \includegraphics[width=\textwidth]{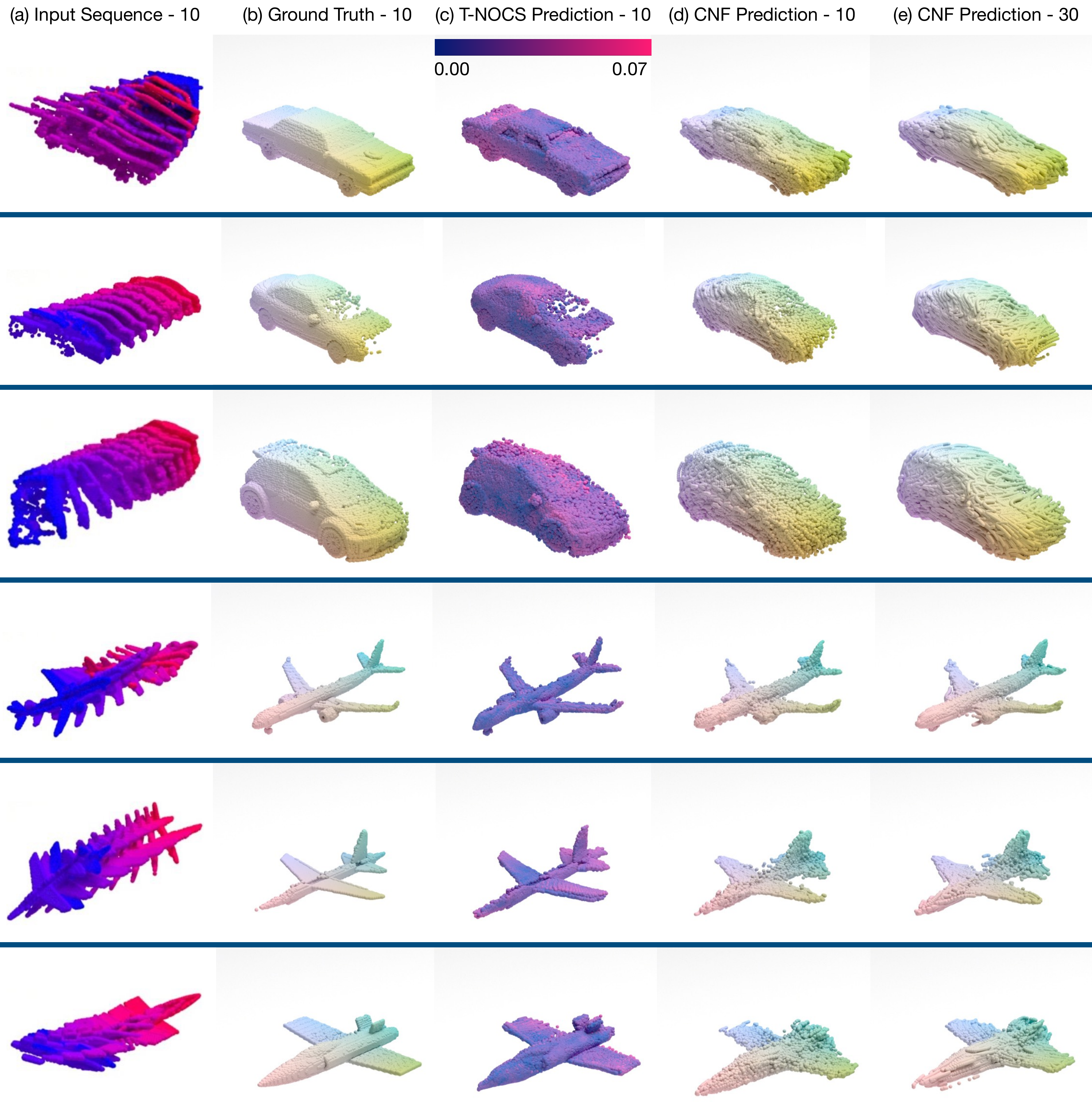}\vspace{-2mm}
    \caption{Canonicalization, aggregation, and dense reconstruction of rigid motion sequences by the full CaSPR model. Each sequence shows (a) the 10 observed raw partial point cloud frames given as input to CaSPR, (b) the GT partial reconstruction based on the observed frames, (c) the partial reconstruction achieved by aggregating T-NOCS predictions from TPointNet++ with color mapped to spatial error, (d) the aggregated prediction after reconstructing the 10 observed frames with the CNF, and (e) the aggregated prediction when interpolating 30 frames using the CNF.}
    \label{appendix:fig:aggregationresults}
    \vspace{-\baselineskip}
\end{figure*}
\stepcounter{appendixfigure}
\begin{figure*}[t!]
    \centering
    \includegraphics[width=\textwidth]{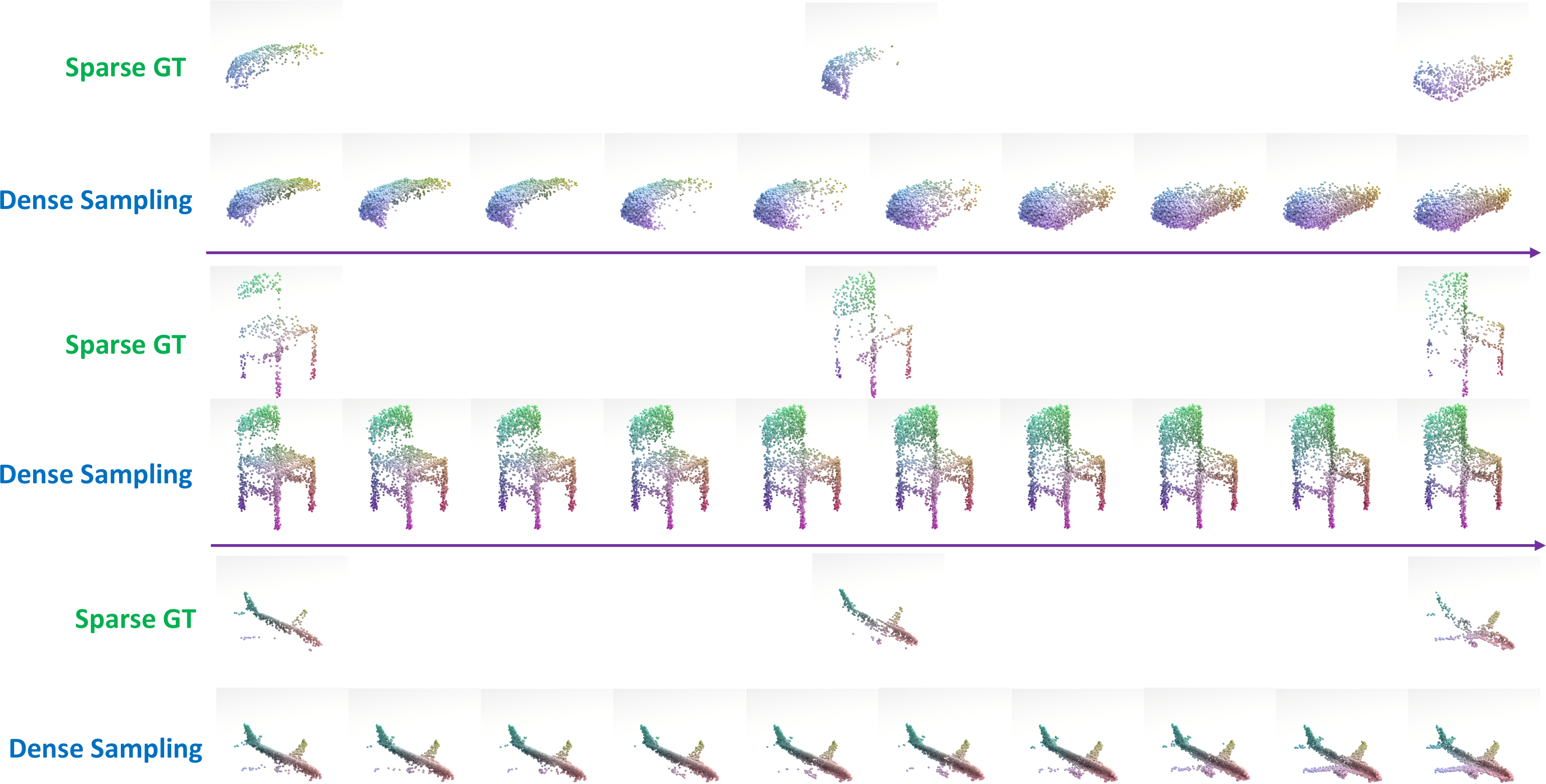}\vspace{-2mm}
    \caption{Examples of spatiotemporal interpolation to reconstruct sparse, partial input sequences. The sparse GT canonical point cloud for each sequence is shown in the top row; the dense CaSPR reconstruction using the CNF is shown in the bottom row.}
    \label{appendix:fig:rigidinterp}
    \vspace{-\baselineskip}
\end{figure*}

The \emph{PointFlow} baseline operates on \textbf{single already-canonical partial point cloud frames}, while CaSPR and \emph{CaSPR-Atlas} take in raw world-space sequences of partial point clouds. To reconstruct a sequence, \emph{PointFlow} can easily reconstruct the observed (canonical) frames by simply autoencoding each frame independently. However, to allow reconstruction of intermediate unobserved steps, we must use linear interpolation in the shape feature space from surrounding observed frames, as described in the main paper.

Median reconstruction errors are presented in Tab. 2 of the main paper. Mean and standard deviation are shown here in~\cref{appendix:table:reconmeans}. Generally, the CaSPR variants have a higher standard deviation than PointFlow. This is likely because CaSPR methods must canonicalize the input in addition to reconstructing it, so any errors in this first step may compound in the reconstruction causing some occasional high errors. A qualitative comparison is shown in~\cref{appendix:fig:pointflowatlascompare}. The \emph{CaSPR-Atlas} baseline has perhaps deceivingly poor EMD errors. As discussed in the main paper, the patch-based approach has difficulty reconstructing the true point distribution of the partial view and may cause some areas to be much more dense or sparse than they should (see chairs in~\cref{appendix:fig:pointflowatlascompare}). Because EMD requires a bijection, these overly dense areas are paired with distant points causing large errors. However, qualitative and CD results suggest the approach has some advantages: the reconstructed point cloud tends to be less noisy and capture local detail better than its CNF-based counterparts.

Additional results of the full CaSPR model reconstructing 10-frame input sequences of rigid, partial point clouds are shown in~\cref{appendix:fig:aggregationresults}. Please see the caption for details. Note that the shown T-NOCS predictions are using TPointNet++ trained jointly within the full CaSPR model rather than individually as in the ``Canonicalization" experiments.

\paragraph{Rigid Spatiotemporal Interpolation}
Additional results of the full CaSPR architecture reconstructing a sparse, partial input sequence are shown in~\cref{appendix:fig:rigidinterp}. In each sequence, the model is given 3 frames with 512 points (with GT canonical point cloud shown as \emph{Sparse GT}) and reconstructs any number of densely sampled steps (10 are shown as \emph{Dense Sampling}, each with 2048 points).

\stepcounter{appendixtable}
\begin{wraptable}[9]{r}{0.5\textwidth}
\vspace{-26pt}
\hspace{-4mm}
\caption{\small Pose estimation performance showing mean and (standard deviation). Supplements Tab. 3 in the main paper.}
\begin{center}
\scalebox{0.65}{
\setlength{\tabcolsep}{3pt}
\hspace{-4mm}
\begin{tabular}{ l c c c c }
\toprule
\textbf{Method} & \textbf{Category} & \textbf{Trans Err} & \textbf{Rot Err}($^\circ$) & \textbf{Point Err} \\
\midrule
RPM-Net & Cars & \textbf{0.0071}	(0.0102) &	\textbf{2.1677} (10.0952) & \textbf{0.0087} (0.0146)  \\
CaSPR &  & 0.0116 (0.0245) & 4.9597 (22.8311) & 0.0203 (0.0645)  \\
\midrule
RPM-Net & Chairs & \textbf{0.0029} (0.0031) & \textbf{0.6212} (3.3530) & \textbf{0.0042} (0.0078)  \\
CaSPR & & 0.0094 (0.0127) & 3.0264 (9.5897) & 0.0152 (0.0367) \\
\midrule
RPM-Net & Airplanes & \textbf{0.0050} (0.0076) & \textbf{2.2703} (16.2945) & \textbf{0.0070} (0.0190)  \\
CaSPR &  & 0.0083 (0.0144) & 3.6740 (16.9152) & 0.0144 (0.0456) \\
\bottomrule
\end{tabular}}
\end{center}
\label{appendix:table:posemeans}
\end{wraptable}

\paragraph{Rigid Pose Estimation}
We solve for object pose in a post-processing step that leverages the world-canonical correspondences given by the output of TPointNet++. We use the full TPointNet++ architecture trained as in the ``Canonicalization" evaluation above. For each frame independently, we run RANSAC~\cite{fischler1981ransac} using 4 points to perform the fitting and with an inlier threshold of $0.015$.

We compare our approach to a recent method for robust rigid registration called RPM-Net~\cite{yew2020rpmnet}. This is an algorithm specially designed for pairwise point cloud registration that iteratively estimates the transformation parameters of possibly-partial point clouds by estimating soft correspondences in the inferred feature space. Because this method can only operate on pairs of point clouds, during training we give it the raw partial point cloud (1024 points) along with the corresponding GT canonical point cloud (1024 points with permuted ordering) as input. This contrasts with TPointNet++ that only receives the raw partial point cloud at each step, and must \emph{predict} the canonical point cloud to establish correspondences. At test-time, we instead use 2048 points for the input point clouds, and the raw points and GT canonical points are randomly sampled so they are not in perfect correspondence.

\stepcounter{appendixfigure}
\begin{figure*}[t!]
    \centering
    \includegraphics[width=0.85\textwidth]{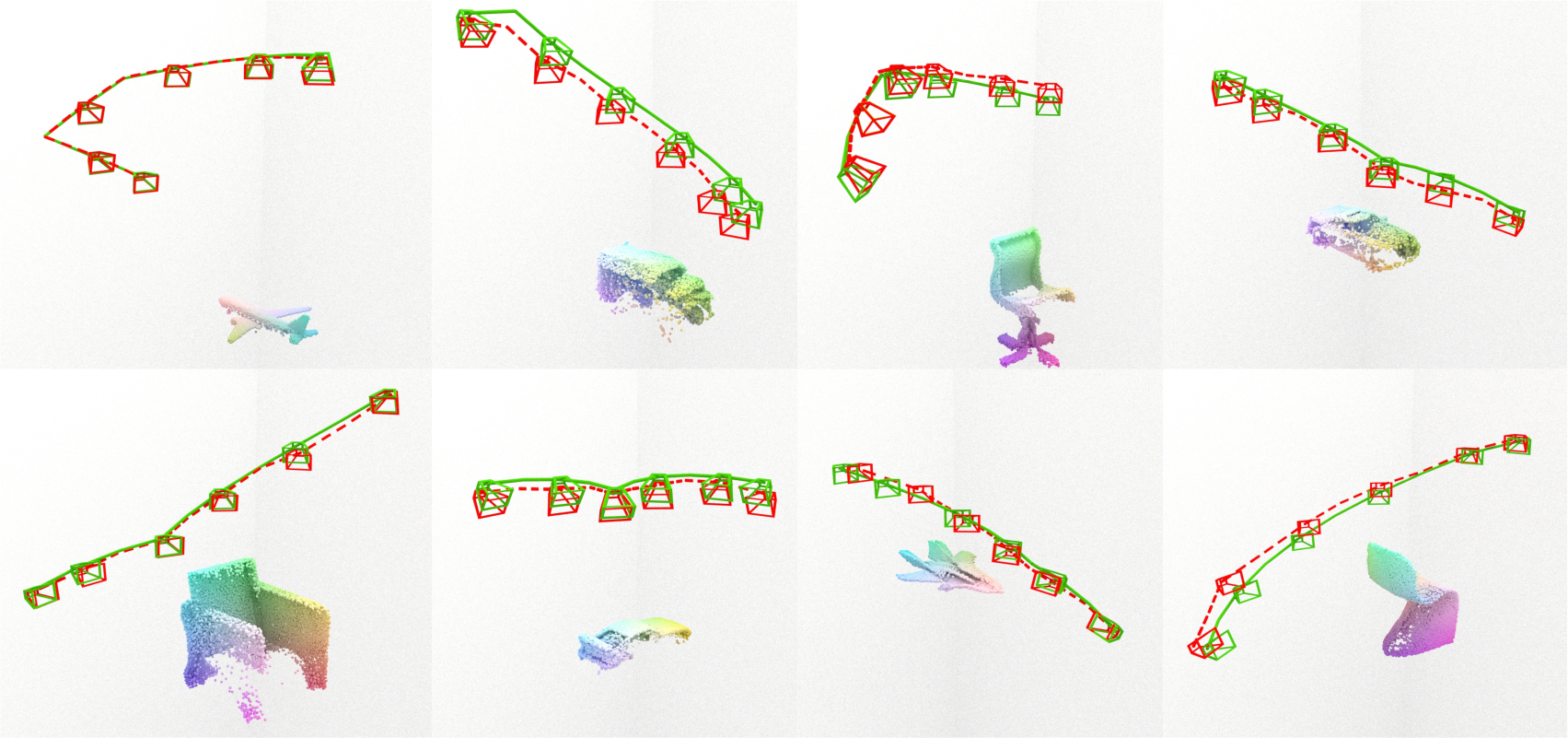}\vspace{-2mm}
    \caption{Additional camera pose estimation results. Ground truth trajectories are shown in solid green and the CaSPR prediction in dashed red.}
    \label{appendix:fig:campose}
    \vspace{-2mm}
\end{figure*}
\stepcounter{appendixfigure}
\begin{figure*}[t!]
    \centering
    \includegraphics[width=0.9\textwidth]{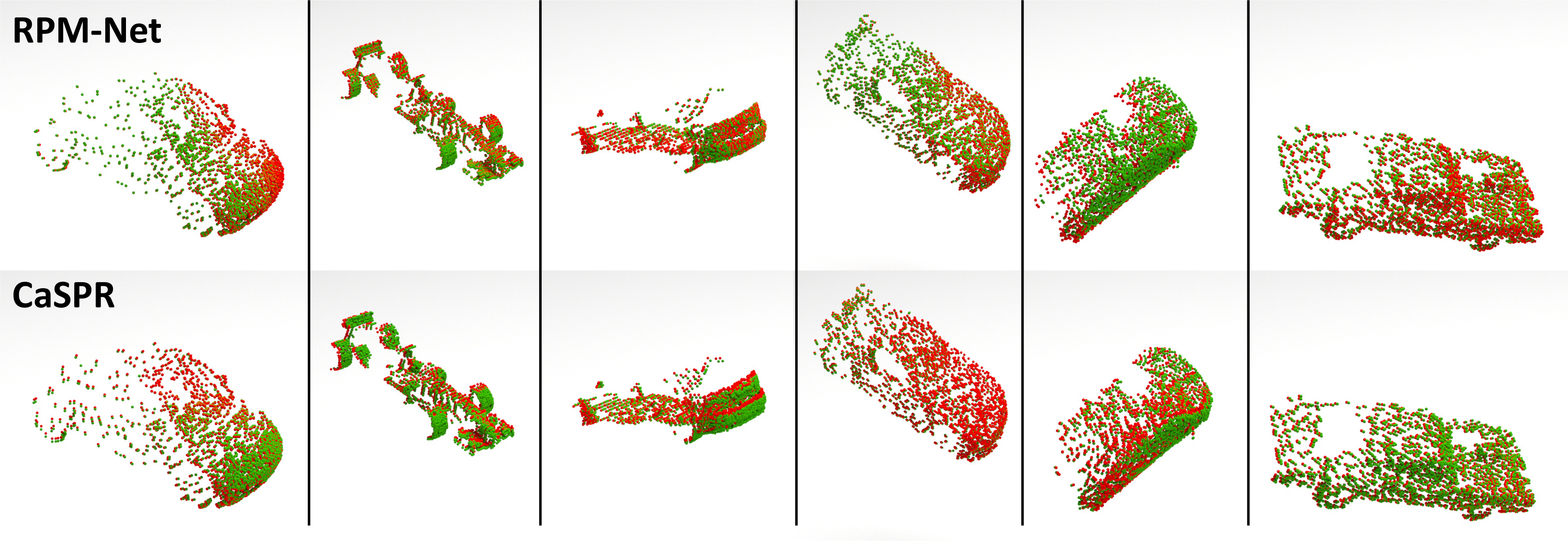}\vspace{-2mm}
    \caption{Rigid pose estimation comparison. Each column shows a frame from a different partial point cloud sequence. Predicted object pose (red points) is shown compared to the GT depth point cloud (green points) for each method. Both methods are very accurate. }
    \label{appendix:fig:rpmcompare}
    \vspace{-\baselineskip}
\end{figure*}

Median errors appear in Tab. 3 of the main paper, but mean and standard deviation results are shown here in~\cref{appendix:table:posemeans}. Though TPointNet++ does not outperform RPM-Net on any shape category, the minimal gap in performance is impressive considering that TPointNet++ has to solve a much harder task (\textit{pose estimation}) than RPM-Net, which receives both the world and canonical point clouds as input and iteratively solves the simpler \textit{pairwise registration} task. As seen in~\cref{appendix:fig:rpmcompare}, the qualitative difference between the two methods is nearly imperceivable. Additional qualitative camera pose estimation results from CaSPR are shown in~\cref{appendix:fig:campose}.

\paragraph{Non-Rigid Reconstruction and Temporal Correspondences}
We compare CaSPR to Occupancy Flow (OFlow)~\cite{niemeyer2019occupancy} on the task of reconstructing Warping Cars sequences and estimating correspondences over time. Because OFlow uses an implicit occupancy shape representation, this dataset contains complete shapes with a clearly defined inside and out. The OFlow baseline uses the point cloud completion version of the model, which leverages a PointNet-ResNet architecture for both the spatial and temporal encoders. OFlow is trained with the reconstruction loss only (\ie~it does not explicitly use a correspondence loss).

\stepcounter{appendixtable}
\begin{wraptable}{r}{0.6\textwidth}
\vspace{-33pt}
\caption{\small Deformable reconstruction and correspondences mean and (standard deviation). Supplements Tab. 4 in the main paper. }
\begin{center}
    \scalebox{0.7}{
    \begin{tabular}{ l c c | c c }
        \toprule
         & \multicolumn{2}{c}{\emph{Reconstruction}} & \multicolumn{2}{c}{\emph{Correspondences}} \\
         \textbf{Method} & \textbf{CD} & \multicolumn{1}{c}{\textbf{EMD}} & \textbf{Dist} $t_1$ & \textbf{Dist} $t_{10}$ \\
        \midrule
        OFlow &  1.764 (0.913) & 24.247 (14.74) & \textbf{0.011} (0.003) & \textbf{0.032} (0.007) \\
        CaSPR  & \textbf{0.992} (0.256) & \textbf{12.864} (5.856) & 0.014 (0.002)	& 0.037 (0.009) \\
\bottomrule
\end{tabular}}
\end{center}
\vspace{-6mm}
\label{appendix:table:deformmeans}
\end{wraptable}

Both methods are trained on sequences of 10 frames with 512 points, and tested on sequences of 10 frames with 2048 points. Due to restrictions of the OFlow encoder, the points at each frame in the input sequence are in correspondence over time (note that this is \textbf{not} a requirement for CaSPR, which can accurately estimate temporal correspondence even if this is not the case as in most real-world applications) and we must use the same number of timesteps at training and test time. To reconstruct a sequence with OFlow, we reconstruct the mesh at the first time step based on the occupancy network predictions, then randomly sample 2048 points on this mesh and advect them forward in time with the predicted flow field. For CaSPR, we advect the latent feature forward in time to each desired timestep as per the usual, then reconstruct each frame with the CNF using the \emph{same} Gaussian samples at each step to achieve temporal continuity. 

Median reconstruction and correspondence errors are reported in Tab. 4 of the main paper. Here we show the mean and standard deviations in~\cref{appendix:table:deformmeans}. Reconstruction errors are measured at all 10 observed timesteps by randomly sampling 2048 ground truth points, while correspondence errors are measured at only the first and last steps using the procedure detailed in the main paper. Additional qualitative results are shown in~\cref{appendix:fig:deformresults}.

\stepcounter{appendixfigure}
\begin{figure*}[t!]
    \centering
    \includegraphics[width=\textwidth]{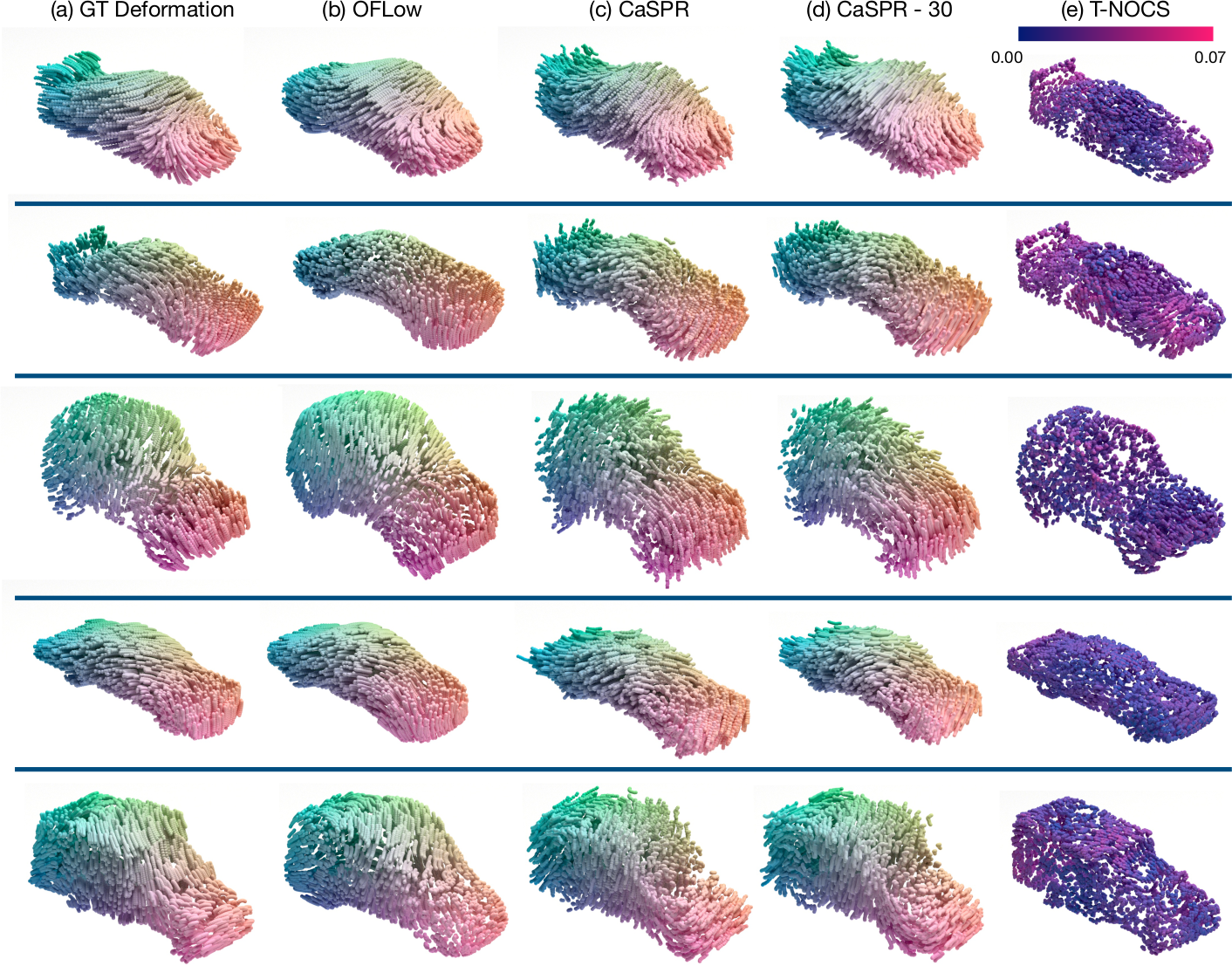}\vspace{-2mm}
    \caption{Reconstruction results on Warping Cars data. Each sequence is 10 steps in length and we show point trajectories over time for (a) the ground truth input sequence, (b) the reconstruction from Occupancy Flow, (c) the reconstruction at the 10 observed steps with CaSPR, (d) 30 interpolated steps with CaSPR, and (e) the T-NOCS prediction from TPointNet++. }
    \label{appendix:fig:deformresults}
    \vspace{-\baselineskip}
\end{figure*}


\end{document}